\documentclass[10pt,twocolumn,letterpaper]{article}

\usepackage{longtable, booktabs}
\usepackage[pagenumbers]{cvpr} 
\usepackage{algorithm}
\usepackage{algpseudocode}

\definecolor{cvprblue}{rgb}{0.21,0.49,0.74}
\usepackage[pagebackref,breaklinks,colorlinks,allcolors=cvprblue]{hyperref}

\def\paperID{21026} 
\def\confName{CVPR}
\def\confYear{2026}

\title{GaussianVision: Vision-Language Alignment from Compressed Image Representations using 2D Gaussian Splatting}

\author{
\begin{tabular}{cccc}
Yasmine Omri & Connor Ding & Tsachy Weissman\textsuperscript{*} & Thierry Tambe\textsuperscript{*}\\
{\tt\small yomri@stanford.edu} & {\tt\small czsding@stanford.edu}  & 
{\tt\small tsachy@stanford.edu} & {\tt\small ttambe@stanford.edu}
\end{tabular}
\\
\\[1ex]
\textsuperscript{*}Equal advising 
\\
Department of Electrical Engineering, Stanford University
}

\begin{document}

\twocolumn[{%
  \renewcommand\twocolumn[1][]{#1}%
  \vspace{-1.25cm}
  \maketitle
  \vspace{-1cm}

  \begin{center}
  \hspace{-0.5cm}
    \includegraphics[width=0.99\textwidth]{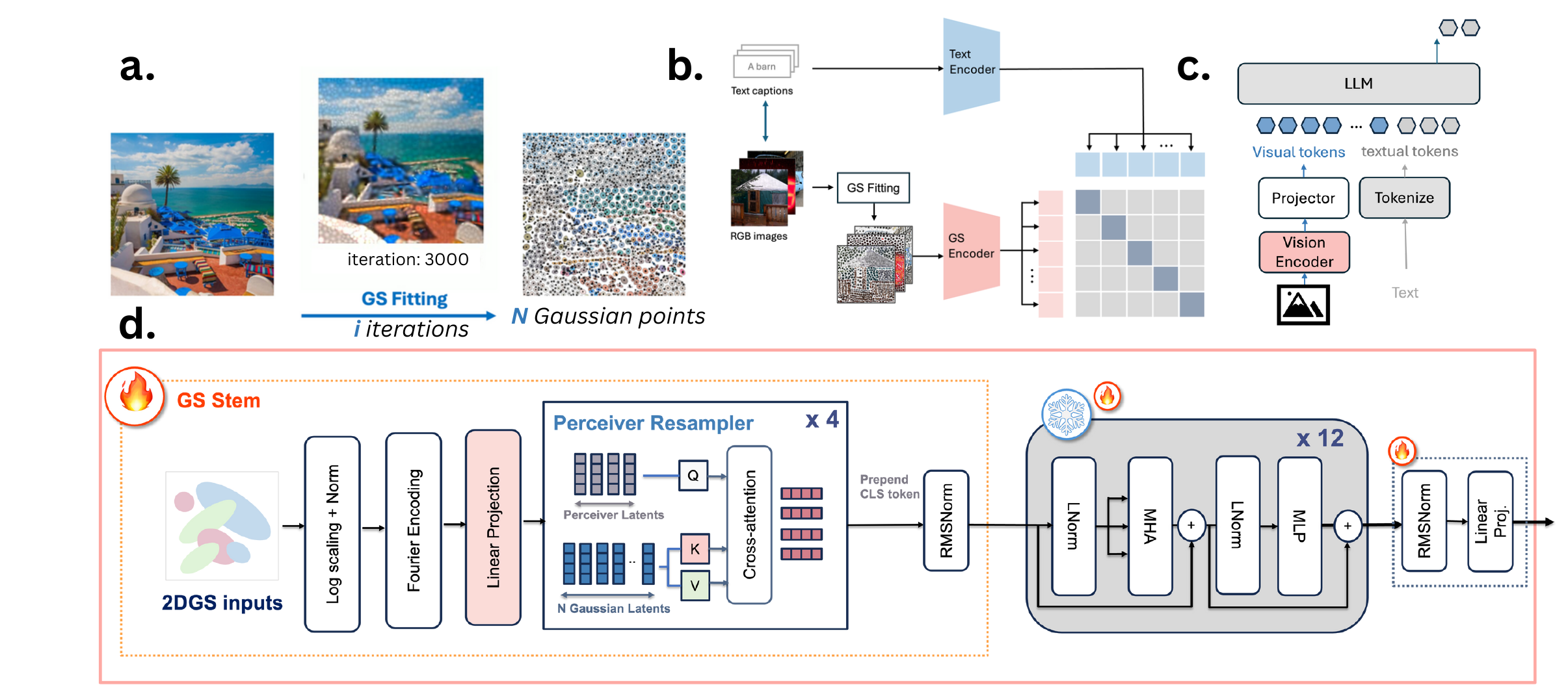}

    \captionsetup{type=figure}
    \caption{\textbf{(a)} Illustration of 2D Gaussian Splatting (2DGS) for image fitting. Each image is represented as a sparse mixture of anisotropic Gaussians parameterized by position, covariance, and color. Summing contributions from all splats reconstructs the original image (with a minor and configurable degradation loss), enabling compact, spatially adaptive representations. \textbf{(b)} 2DGS adaptation of contrastive language-image pre-training (CLIP). \textbf{(c)} Architecture of an autoregressive visual language model (VLM). \textbf{(d)} Architecture of our 2DGS-adapted CLIP pipeline: a splat-aware stem embeds a configurable number of Gaussian points using Fourier features, log scaling, normalization layers, and projections. These embeddings are processed by a majoritarily frozen RGB pretrained transformer.}
    \label{fig:overview}
  \end{center}
}]  

\begin{abstract}
Modern vision–language pipelines are driven by RGB vision encoders trained on massive image–text corpora. While these pipelines have enabled impressive zero-shot capabilities and strong transfer across tasks, they still inherit two structural inefficiencies from the pixel domain: (i) transmitting dense RGB images from edge devices to the cloud is energy-intensive and costly, and (ii) patch-based tokenization explodes sequence length, stressing attention budgets and context limits. We explore 2D Gaussian Splatting (2DGS) as an alternative visual substrate for alignment: a compact, spatially adaptive representation that parameterizes images by a set of colored anisotropic Gaussians. We develop a scalable 2DGS pipeline with structured initialization, luminance-aware pruning, and batched CUDA kernels, achieving over $90\times$ faster fitting and $\sim97\%$ GPU utilization compared to prior implementations. We further adapt contrastive language-image pre-training (CLIP) to 2DGS by reusing a frozen RGB-based transformer backbone with a lightweight splat-aware input stem and a perceiver resampler, training only $\sim9.7 - 13.8\%$ of the total parameters.
On a 12.8M dataset from DataComp, GS encoders yield competitive zero-shot performance on 38 datasets from the CLIP benchmark while compressing inputs $3$–$23.5\times$ relative to pixels. Our results establish 2DGS as a viable multimodal substrate, pinpoint architectural bottlenecks, and open a path toward representations that are \textit{both} semantically powerful and transmission-efficient for edge–cloud learning. 
\end{abstract}

\section{Introduction}
\label{sec:intro}

Expressive yet compact visual representations are central to progress in computer vision and multimodal learning. Recent advances in large multimodal models (LMMs), particularly visual language models (VLMs), have extended the reasoning capabilities of large language models beyond text, enabling impressive capabilities in visual question answering, image captioning, and cross-modal understanding. VLMs leverage a shared embedding space to enable joint reasoning over text, images, and other modalities \cite{Liu2023ImprovedBaselines, Li2024LLaVAOneVision, Li2023BLIP2, Lin2024VILA, AdeptAI2023Adept, Liu2024LLaVANext}, as illustrated by Fig. 1c. Central to this paradigm are vision encoders that map raw images into sequences of visual tokens, semantically aligned with textual embeddings. The canonical pipeline starts from (often high-resolution) RGB images, which are patchified using a uniform grid and routinely projected into hundreds or thousands of tokens. These encoders typically adopt a vision transformer (ViT) backbone, trained via contrastive language–image pre-training (CLIP) on large image–caption datasets \cite{Radford2021CLIP}. While effective, this paradigm suffers from two systemic inefficiencies. (1) Reliance on high-density RGB inputs imposes subtantial data-transfer and energy costs in edge–cloud deployments: even when compressed into formats such as JPEG for transfer, visual data remains bandwidth-intensive and requires costly decoding before processing. (2) Patch-based tokenization causes a severe token explosion: whereas a text prompt may require only a few dozen tokens, vision encoders routinely emit hundreds or thousands per image, many of which are redundant. This redundancy inflates memory and compute costs due to the quadratic scaling of attention, threatening the efficiency and sustainability of large-scale multimodal training and deployment. 
We hypothesize that part of this inefficiency stems from a mismatch between human-centric visual formats and the needs of machine learning systems. RGB pixel arrays are tailored for human perception, with dense spatial correlations, but may include far more detail than models require. Unlike humans, who rely on full-resolution imagery, learning systems could benefit from more abstract representations that emphasize semantic and structural content while discarding perceptually oriented redundancy. This perspective raises a central research question: instead of transmitting perceptually complete images intended for human viewing, \textbf{can we design compact, semantically rich representations that preserve information critical for downstream learning while substantially reducing transmission overhead?}





\vspace{0.5cm}
This work presents the first large-scale exploration of 2D Gaussian Splatting (2DGS) as a visual representation for vision–language alignment in multimodal systems. We ask a central question: \textbf{Can 2D Gaussian splats serve as an efficient and effective intermediate representation for generating text-aligned embeddings in contrastive vision–language models?} To address this, we explore both system-level optimizations for scalable 2DGS and architectural adaptations of contrastive language–image pre-training. Our contributions are two-fold:

\begin{itemize}
\item \textbf{\textit{Systems and Algorithmic Optimizations for Scalable 2DGS:}} We introduce structured initialization, luminance-aware L1 pruning, and batched CUDA kernels, achieving over $90\times$ faster fitting (compared to existing baselines) and $97\%$ GPU utilization. These optimizations make it feasible to generate millions of GS-encoded images for large-scale training.
\item \textbf{\textit{Large-Scale CLIP Adaptation for Gaussian Splat Inputs:}} We systematically adapt the CLIP framework to operate on 2DGS representations, via a novel GS stem and a lightweight two-stage framework for efficient transfer learning from established vision encoders.
\end{itemize}

 We position this work not as an incremental patch-based encoder improvement, but as a first step toward a fundamentally different paradigm: building multimodal systems from inherently compact, adaptive representations rather than post-hoc compression of tokens.


\section{Motivation and Related Work}
\label{sec:related_work}
 \hspace*{2em}\textit{\textbf{The Edge-Cloud Data Transfer Challenge}} 
 Modern multimodal pipelines increasingly depend on edge–cloud frameworks, where bandwidth-limited devices capture high-resolution visual data and transmit it to cloud-based vision encoders for semantic processing. Yet, this paradigm hides a critical inefficiency: transmitting dense RGB-based visual data remains energy-intensive, particularly for resource-constrained edge devices. Empirical studies indicate that mobile uplink transmission (the primary mode for edge devices) alone consumes approximately 0.1–0.2 kWh per GB, (as compared to $\sim$0.03 kWh per GB for fixed broadband connections) \cite{EU_ICT_Impact_2020}. At scale, this may translate to tens of kilowatt-hours for just a few gigabytes of visual data. For instance, a single hour of 1080p video transmission consumes 0.38-0.68 kWh over mobile networks, energetically equivalent to powering a typical home for hours \cite{GSMA_2025_mobile_data_energy}. As such, the energy and infrastructure costs of naive RGB data transfer can quickly dominate operational budgets in real-world multimodal systems.
\\ 
\hspace*{2em}\textit{\textbf{The Token Explosion Problem and the Increasing Strain on Context Lengths}} 
The second challenge emerges in downstream processing, where current vision encoders generate sequences of hundreds or thousands of tokens per image. For example, LLaVA-1.5 \cite{Liu2023ImprovedBaselines} emits 576 tokens for a $336\times336$ image, while LLaVA-NeXT \cite{Liu2024LLaVANext} outputs over 2,880 tokens for $672\times672$ inputs, compared to about 20–50 tokens for a typical sentence in a text prompt. This visual information density creates severe computational bottlenecks, as self-attention scales quadratically with sequence length. 
To address this, recent work has focused on post-hoc token reduction. PruMerge \cite{Shang2024prumerge} and VisionZip \cite{Yang2024visionzip} implement text-agnostic selection strategies, with the former requiring LLM-level fine-tuning and the latter projector-level fine-tuning for optimal performance. ToMe \cite{tome} integrates token merging directly into encoder blocks via cosine-similarity–based selection. Progressive sparsification methods such as FastV \cite{Chen2024fastv} and SparseVLM \cite{Zhang2025sparsevlm} prune tokens layer by layer, guided by attention weights or cross-modal relevance. More recently, cluster-based aggregation \cite{Omri2025tokens} has shown that simple, finetuning-free grouping can outperform prior methods, with minimal loss in downstream performance. Strikingly, discarding up to 89\% of visual tokens often has little effect on VQA benchmarks \cite{Li2023POPE, hudson2019gqa, singh2019textvqa, Lu2022ScienceQA, Liu2024MMBench, Fu2023MME, Yu2024MMVet, gurari2018vizwiz}, and in some cases even improves accuracy. Even naive strategies like spatial or random sampling perform competitively, highlighting the vast redundancy of current visual tokens. While effective in reducing sequence length, these approaches remain constrained by patch-based tokenization; they are, in essence, band-aid solutions to an architectural inefficiency. The fact that most tokens can be discarded without harming performance underscores a deeper issue: today’s encoders may be producing large volumes of redundant, semantically shallow information. This not only inflates compute, but also raises a broader question: are patch-based encoders truly the right substrate for multimodal alignment? These findings suggest an opportunity to move beyond post-hoc pruning toward representations that are inherently compact, semantically expressive, and computationally efficient.\\ 
\\ 
\hspace*{2em}\textit{\textbf{Compressed and Alternative Image Representations}}
The limitation of post-hoc token reduction extends beyond compute efficiency. Such methods assume that optimal visual representations must originate from dense pixel grids, and only later compress them. This paradigm overlooks the possibility that alternative input representations could be inherently more compact, expressive, and better aligned with the statistical patterns neural networks exploit. Traditional codecs such as JPEG achieve high compression ratios but yield representations that lack interpretability for learning systems, though some work has explored making compressed formats more machine-friendly. Implicit neural representations (INRs) instead encode signals as continuous functions parameterized by neural networks, achieving high fidelity in compression and reconstruction \cite{sitzmann2020SIREN}. Recent studies have even treated INR parameters as input embeddings for downstream tasks, with promising results \cite{Bauer2023SpatialFuncta, Dupont2022FromDataToFuncta}. However, INRs remain computationally costly to optimize and are ill-suited for large-scale training pipelines. A recent alternative is 2D Gaussian Splatting (2DGS): a 2D adaptation \cite{Zhang2024GaussianImage} of 3D Gaussian Splatting for real-time radiance field rendering \cite{Kerbl2023ThreeDGaussianSplatting}.
\section{2D Gaussian Splatting Preliminaries}
\label{sec:background}


Building on 3D Gaussian Splatting (3DGS) for real-time novel-view synthesis \citep{Kerbl2023ThreeDGaussianSplatting}, \cite{Zhang2024GaussianImage} introduced GaussianImage, adapting the concept for single-image compression. Instead of heavy 3D Gaussians requiring tens of parameters, GaussianImage uses compact 2D Gaussians parameterized by only eight values: 2D position, anisotropic covariance, and color, as illustrated by Fig. 1a. Images are reconstructed by summing contributions from all Gaussians, following Equation \ref{eq:gs_reconstruction}.


\vspace{-0.8cm}
{\small
\begin{equation}
\begin{split}
\hat{I}\!\left(x,y;\,\{\boldsymbol{\mu}_i,\boldsymbol{\Sigma}_i,\mathbf{c}_i\}_{i=1}^{n}\right)
&= \sum_{i=1}^{n} \mathbf{c}_i\,
\exp\!\Bigg\{-\tfrac{1}{2}
\left(\begin{bmatrix}x\\y\end{bmatrix}-\boldsymbol{\mu}_i\right)^{\!\top} \\
&\qquad \boldsymbol{\Sigma}_i^{-1}
\left(\begin{bmatrix}x\\y\end{bmatrix}-\boldsymbol{\mu}_i\right)
\Bigg\},
\end{split}
\label{eq:gs_reconstruction}
\end{equation}
}
\vspace{-0.6cm}



where $\hat{I}(x,y)$ is the predicted image intensity at pixel $(x,y)$, $\mu_i\in\mathbb{R}^2$ is the Gaussian center, $\Sigma_i\in\mathbb{R}^{2\times2}$ its covariance (often expressed by three free parameters), and $c_i$ its color. 


2D Gaussian Splatting eliminates the need for depth sorting, enabling extremely fast rendering ($1{,}500$–$2{,}000$ FPS, or equivalently over 3X the decoding speed of JPEG), substantial memory savings, and $\sim 5\times$ faster training compared to implicit neural representations (INRs) such as WIRE \cite{saragadam2023wirewaveletimplicitneural} or I-NGP \cite{M_ller_2022} \cite{Zhang2024GaussianImage}.

For further compression, GaussianImage employs a quantization pipeline (16-bit floats for positions, 6-bit integer covariance quantization, residual vector quantization for color, and partial bits-back coding). This achieves rate–distortion performance comparable to INR-based codecs while offering orders-of-magnitude faster decoding \citep{Zhang2024GaussianImage}. Following this, \cite{Zhu2025LIG} proposed LIG (Large Images are Gaussians), a hierarchical two-level scheme that first fits low-frequency content and then refines high-frequency details, significantly improving scalability to ultra-high-resolution images. Extending to video, GSVC \cite{Wang2025GSVC} predicts Gaussians sequentially across frames, prunes low-impact ones, adds new splats as needed, and reinitializes at keyframes, achieving AV1/VVC-level rate–distortion performance while decoding 1080p video at ~1,500 FPS.

\vspace{0.4cm}
\textbf{Beyond compression, 2DGS has been explored as a learnable substrate for vision tasks.} GViT \cite{Hernandez2025GViT} replaces pixel patches with hundreds of Gaussians, and jointly optimizes them with a Vision Transformer. A differentiable renderer enforces image reconstruction, while classification gradients push Gaussians toward discriminative regions, yielding a compact, interpretable representation that matches ViT-B performance on ImageNet-1K (76.9\% top-1) supervised classification. \cite{Dong2025GaussianToken} proposes GaussianToken, which encodes images as Gaussians with continuous spatial parameters and feature coefficients, quantized via a VQ codebook and concatenated with spatial attributes. This hybrid token design improves reconstruction fidelity over VQ-VAE tokenizers. Yet, existing efforts remain limited to small-scale settings. Our work addresses this gap, providing the first large-scale study of 2DGS within vision–language pre-training, asking whether its promising properties can support vision–language alignment at scale.
\section{Optimizations for Efficient and Scalable 2D Gaussian Splatting}
\label{sec:optimization}

\subsection{Challenges}
Large-scale vision–language alignment requires fitting millions of images, but existing 2DGS pipelines are still too slow, leaving substantial headroom in batching, memory layout, and GPU utilization. As a result, naïvely applying 2DGS at scale is infeasible without both algorithmic refinements and systems-level restructuring. We therefore begin our study by developing a set of optimizations that make high-throughput 2DGS fitting practical at the scale required for downstream alignment.

\subsection{Specialized CUDA Kernels for Scalable 2DGS}

Our implementation builds on the open-source GaussianImage codebase \cite{Zhang2024GaussianImage}, which first introduced 2D Gaussian Splatting for high-speed image representation. Subsequent work by \cite{Zhu2025LIG} streamlined this design, restructuring the kernels and data flow specifically for 2D operations and thereby yielding a cleaner, more efficient foundation for image representation. This codebase is structured as follows: the forward pass is decomposed into two phases: (1) a projection stage, wherein each Gaussian primitive is projected onto a 2D grid of pixels. The grid is tiled across thread blocks, where each block corresponds to a 2D tile of the output image. Within a block, individual threads are assigned to pixels, computing Gaussian weights and contributions in parallel. This design ensures that Gaussian evaluation is parallelized across both the spatial dimension (pixels) and across different Gaussians, and (2) a rasterization stage, wherein the projected Gaussians are accumulated to produce the final output image at each iteration. This stage handles blending and weighting of Gaussians for each pixel. Unlike projection, where computation is Gaussian-centric, rasterization is pixel-centric, with threads accumulating contributions into shared output buffers. The backward pass is handled by a dedicated kernel that computes gradients of the loss with respect to Gaussian parameters (positions, covariances, colors). Workload partitioning is inverted relative to the forward pass: each block is assigned to a 2D tile of the input grid, and threads within the block compute pixel-level derivatives. This tiling strategy balances workload across warps and ensures efficient coalesced memory access when accumulating gradients. \\

\begin{figure}[h]
\begin{center}
\includegraphics[width=1\linewidth]{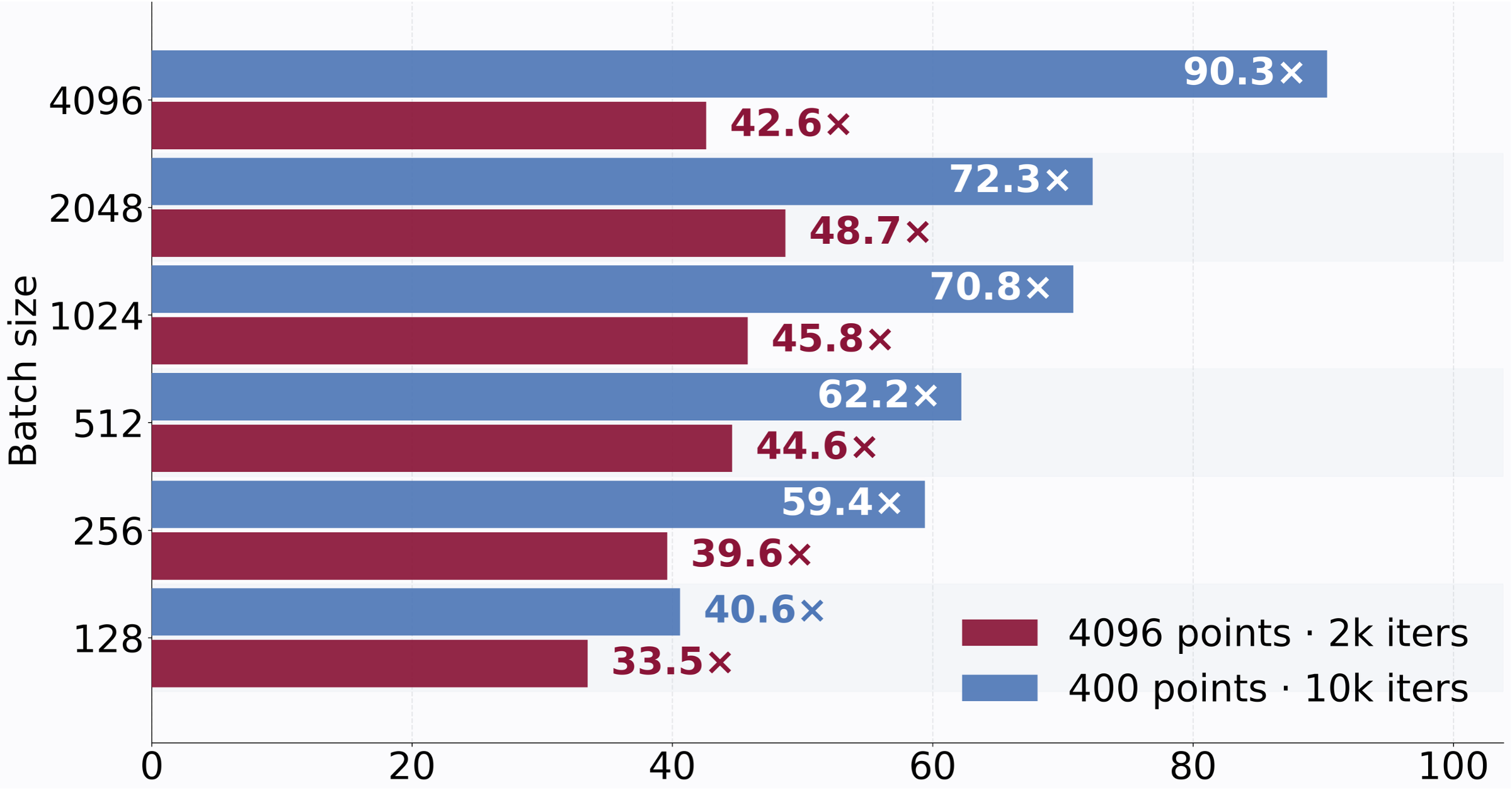} 
\end{center}
\caption{Speedup results achieved by our CUDA kernels compared to the \cite{Zhu2025LIG} baseline, following our batch-aware implementation. Speedups are presented for various batch sizes and Gaussian counts for image resolutions of 224x224. For a batch size of 4096 and 400 Gaussian points per image, we observe a 90.3X speedup compared to the baseline.}
\label{cuda_fig}
\end{figure}

We extend the implementation with batched input support by introducing several key modifications:
\begin{itemize}
    \item Batch-aware memory layout: Indexing and memory access patterns were modified to include the batch dimension, allowing multiple images to be processed concurrently.
    \item Additional synchronization threads were inserted at critical points to ensure that all threads complete their batch-local computations before proceeding, preventing race conditions when accumulating results.
\item Shared memory usage: Shared memory, already used in the original codebase for intermediate accumulation, was expanded to cache per-batch partial results. This reduces the frequency of global memory accesses and improves overall throughput.

\end{itemize}
With these modifications, the codebase now supports high-speed processing of thousands of inputs concurrently, making the implementation suitable for large-scale training pipelines. Notably, our enhanced codebase enables up to 90X speedup in fitting time compared to the baseline open-sourced codebase by \cite{Zhu2025LIG} with ~97\% GPU utilization, profiled using NVIDIA Nsight Compute. We provide additional details on our CUDA optimizations in Appendix~C. Fig. \ref{cuda_fig} presents speedup results for various batch sizes and gaussian counts. Our optimized 2DGS pipeline, including both our algorithmic and system-level optimizations will soon be open-sourced.

\subsection{Structured Initialization}
\begin{figure}[t]
\centering

\begin{tabular}{c c c c c}

\rotatebox{90}{\textcolor{blue}{\small Ours}} &
\hspace{-0.3cm}
\begin{minipage}{0.22\linewidth}
\centering
\includegraphics[width=\linewidth]{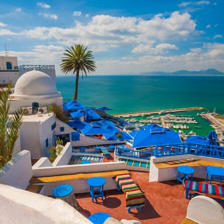}
\\[-1mm]
{\textcolor{blue}{\scriptsize 4900 pts}}
\end{minipage}
&
\hspace{-0.4cm}
\begin{minipage}{0.22\linewidth}
\centering
\includegraphics[width=\linewidth]{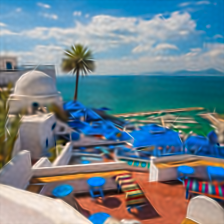}
\\[-1mm]
{\textcolor{blue}{\scriptsize 1600 pts}}
\end{minipage}
&
\hspace{-0.4cm}
\begin{minipage}{0.22\linewidth}
\centering
\includegraphics[width=\linewidth]{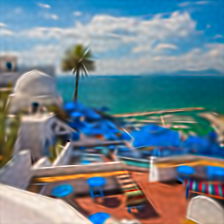}
\\[-1mm]
{\textcolor{blue}{\scriptsize 900 pts}}
\end{minipage}
&
\hspace{-0.4cm}
\begin{minipage}{0.22\linewidth}
\centering
\includegraphics[width=\linewidth]{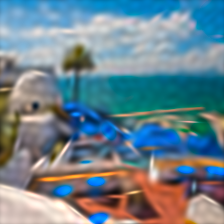}
\\[-1mm]
{\textcolor{blue}{\scriptsize 400 pts}}
\end{minipage}
\\[3mm]

\rotatebox{90}{\textcolor{blue}{\small Random}} &

\hspace{-0.4cm}
\begin{minipage}{0.22\linewidth}
\centering
\includegraphics[width=\linewidth]{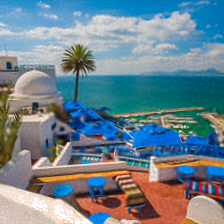}
\\[-1mm]

\end{minipage}
&
\hspace{-0.4cm}
\begin{minipage}{0.22\linewidth}
\centering
\includegraphics[width=\linewidth]{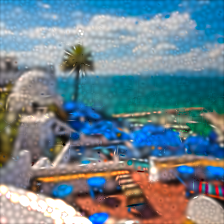}
\\[-1mm]

\end{minipage}
&
\hspace{-0.4cm}
\begin{minipage}{0.22\linewidth}
\centering
\includegraphics[width=\linewidth]{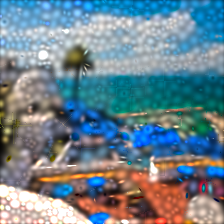}
\\[-1mm]

\end{minipage}
&
\hspace{-0.4cm}
\begin{minipage}{0.22\linewidth}
\centering
\includegraphics[width=\linewidth]{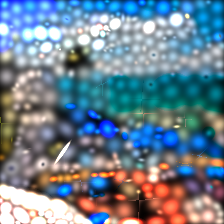}
\\[-1mm]

\end{minipage}

\end{tabular}
\caption{Visualization of reconstruction results for random vs. structured initialization (Ours) for 2DGS fitting for a fixed number of iterations (3000): structured initialization accelerates convergence and achieves higher perceptual quality than random initialization. This is consistent across various compression ratios (ie, numbers of Gaussian points per image) especially for more aggressive compression ratios.}
\label{init_quality}
\end{figure}




Traditional 2D Gaussian splatting approaches follow the conventions established in 3D Gaussian splatting, where random initialization of Gaussian parameters is the standard practice. In 3D scene reconstruction, this random initialization strategy is well-motivated, because the underlying 3D geometry is unknown a priori, and points must be discovered through optimization. The 3D case deals with sparse, incomplete observations of a scene from multiple viewpoints, making a structured initialization impractical without prior geometric knowledge.
However, the 2D image case presents a fundamentally different scenario where we can leverage the pixel-level prior information available in the input image. Unlike 3D reconstruction where we must infer spatial structure, 2D images provide direct access to spatial organization and color information at every pixel location. Our structured initialization strategy exploits this pixel prior through three key components: \\ 
\indent \textbf{(1) Position Initialization:}: Rather than randomly sampling Gaussian centers, we initialize the x,y coordinates of each Gaussian following a uniformly distributed grid pattern across the image dimensions. This grid-based approach maximizes spatial coverage, ensuring that each region receives appropriate representational capacity from the start of optimization.\\ 
\indent \textbf{(2) Covariance Initialization:}: Each Gaussian starts with an isotropic covariance, corresponding to the largest circle that can be fitted into the grid cell.
\\ 
\indent \textbf{(3) Color Initialization}: RGB values are initialized as the average color RGB value of all pixels in the grid cell corresponding to that gaussian. This color inheritance provides a strong starting point for optimization, as each Gaussian begins with a color value that is already locally appropriate for its spatial region.

As illustrated by the example in Fig. \ref{init_quality}, this design provides strong priors that accelerate convergence. With 4,900 Gaussians and 3000 iterations, structured initialization achieves 35.25 PSNR versus 28.24 for random initialization. Even at 400 Gaussians, it maintains 22.04 PSNR versus 17.77, demonstrating both faster convergence and higher asymptotic perceptual quality. Further analysis and experiments are presented in Appendix B.

\subsection{Adaptive Pruning through L1 Regularization and Luminance Thresholding}


\begin{figure}[t]
    \centering

        \includegraphics[width=\linewidth]{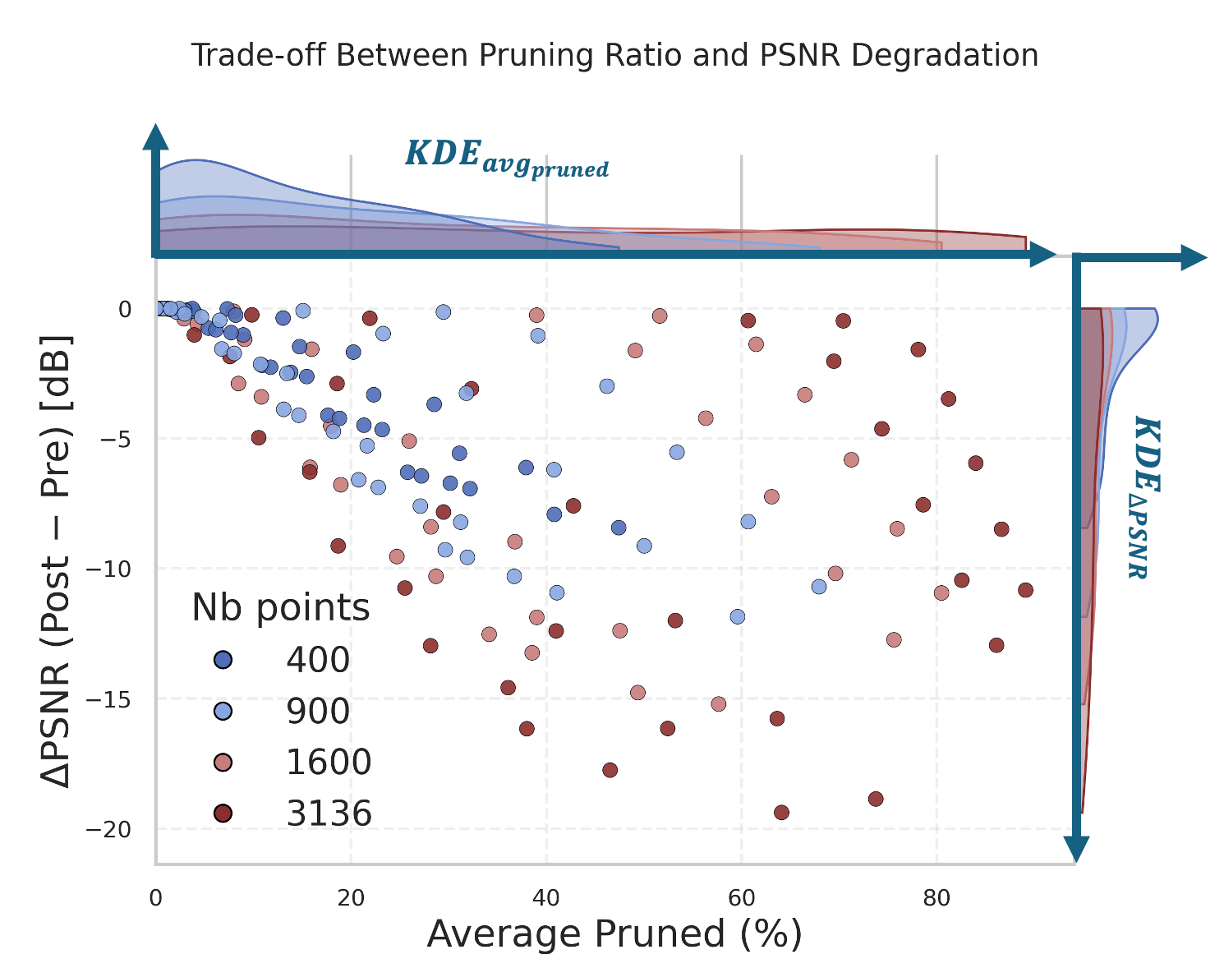}

    \caption{
Trade-off between pruning ratio and reconstruction degradation for different Gaussian budgets (400--3136 points), evaluated over 100 Mini-ImageNet samples per configuration. 
Each marker represents a single hyperparameter setting, while the surrounding shaded KDE envelopes summarize the empirical distribution of $\Delta\mathrm{PSNR}$ for each model size: models with larger initial Gaussian budgets (1600--3136) consistently support higher pruning ratios with minimal loss, while smaller models are more sensitive to sparsification. 
}
    \label{prune_tradeoff}
\end{figure}


Achieving optimal compression while preserving semantic information requires intelligent removal of redundant Gaussian primitives during optimization. Our adaptive pruning strategy combines L1 regularization with luminance-based thresholding to identify and eliminate Gaussians that contribute minimally to overall reconstruction quality.\\ 
\indent\textbf{\textit{L1 Regularization Framework: }} We incorporate L1 penalty terms on the color channels of the Gaussian points during fitting to naturally encourage sparsity during optimization. This regularization framework allows the model to automatically identify low-contribution Gaussians by driving their color parameters toward zero (which effectively reduces their color contribution to pixel reconstructions), making them candidates for removal without explicit supervision.
\[
\mathcal{L}_{\mathrm{GS}}
= 
\frac{1}{B}\sum_{b=1}^{B}
\bigg[
\underbrace{
\frac{1}{HW}\|\hat{\mathbf{I}}^{(b)} - \mathbf{I}^{(b)}\|_2^{2}
}_{\text{L2 reconstruction}}
\;+\;
\lambda_{\mathrm{reg}}
\underbrace{
\big\|\mathbf{C}^{(b)}\big\|_{1}
}_{\text{L1 color reg}}
\bigg],
\]

where 
\(\hat{\mathbf{I}}^{(b)} \in \mathbb{R}^{H \times W \times 3}\) is the rendered Gaussian–splat
prediction for sample \(b\), 
\(\mathbf{I}^{(b)} \in \mathbb{R}^{H \times W \times 3}\) is the corresponding ground--truth
image, 
\(\mathbf{C}^{(b)}\) denotes the concatenated color coefficients of all Gaussians for
sample \(b\), 
\(\|\cdot\|_{2}^{2}\) is the pixel--wise squared \(\ell_2\) error averaged over all
\(H \times W\) pixels, 
\(\|\cdot\|_{1}\) is the \(\ell_1\) norm applied to all color features of the Gaussians,
\(B\) is the batch size, and 
\(\lambda_{\mathrm{reg}}\) is the weighting factor applied to the color regularization
term.

\textbf{\textit{Information-Density Aware Pruning: }} Once the GS fitting is done, we dynamically assess each Gaussian's contribution to overall reconstruction quality against a luminance-based threshold. Gaussians with a low-luminance contribution often correspond to areas with minimal visual information. Our luminance-based thresholding mechanism identifies these regions and marks associated Gaussians for removal, reducing the final number of Gaussian points, while preserving perceptually important visual content.

\[
s_{b,n}
=
\underbrace{
0.2126\,\bigl|R_{b,n}\bigr|
+ 0.7152\,\bigl|G_{b,n}\bigr|
+ 0.0722\,\bigl|B_{b,n}\bigr|
}_{\text{luminance score } \ell(\mathbf{c}_{b,n})}
\]

\[
\mathcal{K}^{(b)}
=
\bigl\{
n \in \{1,\dots,N\}
\,\big|\,
s_{b,n} \ge \tau_{\mathrm{th}}
\bigr\},
\]

\[
\mathbf{c}'_{b,n}
=
\begin{cases}
\mathbf{c}_{b,n}, & n \in \mathcal{K}^{(b)},\\[4pt]
\mathbf{0},       & n \notin \mathcal{K}^{(b)},
\end{cases}
\]

where
\(b \in \{1,\dots,B\}\) indexes images in the batch,
\(n \in \{1,\dots,N\}\) indexes Gaussians within an image,
\(\mathbf{c}_{b,n} = (R_{b,n}, G_{b,n}, B_{b,n}) \in \mathbb{R}^3\) is the RGB vector
of the \(n\)-th Gaussian in sample \(b\),
\(\ell(\mathbf{c}_{b,n})\) is the luminance score defined by the
weighted sum of absolute RGB channels,
\(\tau_{\mathrm{th}} > 0\) is the luminance threshold hyperparameter,
\(\mathcal{K}^{(b)}\) is the index set of Gaussians retained for sample \(b\),
\(\mathbf{c}'_{b,n}\) is the pruned RGB vector after thresholding, and
\(\mathbf{0}\) denotes the zero RGB vector, corresponding to Gaussians
that no longer contribute to the rendered image.


\begin{figure}[htbp]
    \begin{subfigure}[t]{0.23\columnwidth}
        \includegraphics[width=\linewidth]{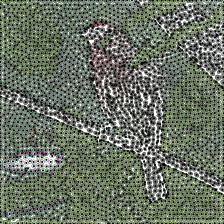}
        \label{fig:row3a}
    \end{subfigure}\hspace{0.5mm}
    \begin{subfigure}[t]{0.23\columnwidth}
        \includegraphics[width=\linewidth]{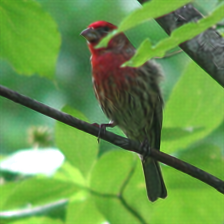}
        \label{fig:row3b}
    \end{subfigure}\hspace{0.5mm}
    \begin{subfigure}[t]{0.23\columnwidth}
        \includegraphics[width=\linewidth]{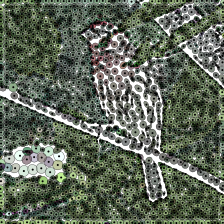}
        \label{fig:row3c}
    \end{subfigure}\hspace{0.5mm}
    \begin{subfigure}[t]{0.23\columnwidth}
        \includegraphics[width=\linewidth]{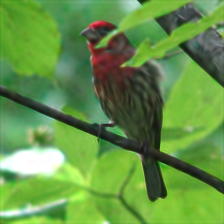}
        \label{fig:row3d}
    \end{subfigure}
\caption{
Visualization of Gaussian splats and reconstructed images for a 3136-point GS fit (2000 iterations). 
Left: $\lambda_{\mathrm{reg}}{=}0$, $\tau_{\mathrm{th}}{=}0$ (0\% pruned: PSNR = 37.43). 
Right: $\lambda_{\mathrm{reg}}{=}10^{-6}$, $\tau_{\mathrm{th}}{=}0.05$ (23.72\% pruned: PSNR = 31.1).
}
\label{prune_demo}
\end{figure}

Fig. \ref{prune_tradeoff} presents a sweep over the luminance-based pruning threshold $\tau_{\mathrm{th}} \in \{0, 0.01, 0.05, 0.10, 0.15, 0.20, 0.25\}$ and the color regularization weight $\lambda_{\mathrm{reg}} \in \{0, 10^{-7}, 5{\times}10^{-7}, 10^{-6}, 5{\times}10^{-6}, 10^{-5}\}$ across four Gaussian point budgets (400, 900, 1600, 3136) and 2000 iterations, averaging each configuration over 100 Mini-ImageNet samples (224x224 each). 
For every setting, we record the resulting sparsity level and the reconstruction change $\Delta\mathrm{PSNR} = \mathrm{PSNR}_{\text{post}} - \mathrm{PSNR}_{\text{pre}}$, capturing the fidelity loss induced by pruning. 
Models initialized with larger Gaussian counts (1600--3136) tolerate aggressive pruning---often $60$--$80\%$---while retaining high reconstruction quality, typically within $2$--$5\,\mathrm{dB}$. An example of this pruning effect is shown in Fig.~\ref{prune_demo}. We further observe that starting with a larger point budget and pruning down typically yields higher final PSNR than allocating a small budget from the outset, demonstrating that redundancy in the initial representation is beneficial for sparsification. We demonstrate this and additional analyses on the interaction between $\lambda_{\mathrm{reg}}$, $\tau_{\mathrm{th}}$, pruning ratio, and experimentation with alternative thresholding strategies in Appendix~B.

\section{Vision–Language Alignment at Scale from 2D Gaussian Splat Representations}
\label{sec:experiments}

\subsection{Experimental Setup}
A central goal of our study is to evaluate whether 2DGS representations can serve as a practical visual substrate for large-scale vision–language models. We leverage the mature inductive biases encoded in RGB ViT backbones, enabling efficient and lightweight adaptation to GS inputs. 
Our RGB reference model is a CLIP ViT-B/16 trained on 12.8M DataComp pairs \cite{Gadre2023DataComp}. To stay within a practical compute budget while retaining a meaningful CLIP baseline, we use a width-512 variant of ViT-B/16 (38M parameters) that we will refer to as  ViT-B/16 (Small), which we verify maintains comparable accuracy to the standard 768-dim version on our setting (Appendix~D). We use the standard Open CLIP hyperparameters \cite{OpenCLIP2021}: learning rate $10^{-3}$, batch size 256, 32 epochs, cosine schedule, and weight decay~0.2. 

We conduct our experiments on the Stanford Marlowe High Performance Computing (HPC) cluster \cite{Kapfer2025Marlowe}, using one node of 8 H100 GPUs.

For our GS encoders, we preprocess the 12.8M images into FP16 Gaussian splat representations with varying budgets of $\{400, 900, 1600, 3136\}$ splats per image. Each splat stores 8 parameters (position, covariance, color), yielding compression ratios of $3\times$–$23.5\times$ relative to RGB pixels. These splats are used as the sole visual input for 2DGS models.
\vspace{-0.2cm}

\subsection{Two-Stage Training Procedure}

\paragraph{Motivation.}
Directly training CLIP from raw 2DGS inputs converges poorly (see Appendix~D). In contrast, aligning 2DGS features to a pretrained RGB embedding space dramatically accelerates convergence. This motivates a two-stage procedure: we first teach the GS stem to match the CLS embeddings of an RGB teacher model, and then perform lightweight multimodal alignment.

\vspace{-0.5cm}
\paragraph{Stage 1: RGB$\rightarrow$GS Distillation.}
This stage involves a GS vision encoder and a frozen RGB vision encoder. We initialize a GS encoder whose transformer backbone shares weights with the RGB ViT-B/16 (Small) baseline. Only the \emph{GS stem}, the module responsible for mapping raw Gaussian parameters to token embeddings, is trained in this stage. The GS stem is illustrated in Fig. 1d, and consists of: (i) log conversion of covariance for outlier reduction (ii) Fourier positional features, (iii) normalization and linear projection to a depth of 128 (iv) a perceiver module with 4 cross-attention layers followed by a projection to the frozen transformer backbone width (512). The total size of the GS encoder matches the RGB teacher (38M). We train this stem for 2 epochs using an MSE loss between L2-normalized CLS embeddings of the RGB teacher and GS student. This aligns the splat representation to the structured geometric manifold learned by the RGB model.
\vspace{-0.5cm}
\paragraph{Stage 2: Parameter-Efficient CLIP Adaptation.}
This stage involves a GS vision encoder and a frozen text encoder. We  perform CLIP contrastive training for 5 epochs, unfreezing only $\sim9.7\%$ of total CLIP parameters: the GS stem, the first two transformer blocks, and the final normalization and projection layers on the vision side. This yields fast convergence at low computational cost. After warmup, we optionally unfreeze a symmetric set of text-side adapter layers (final transformer block, final norm, projection), bringing the total number of trainable parameters to $\sim13.8\%$ and adding a small accuracy gain. 

\subsection{Evaluation Protocol}
We evaluate all models in the zero-shot setting using 38 datasets from the CLIP benchmark \cite{cherti_2025_15403103}, which span natural, fine-grained, synthetic, medical, OCR, and sketch zero-shot classification. Zero-shot evaluation is a standard proxy for vision–language alignment quality and avoids confounding factors from per-dataset fine-tuning.  All experiments are conducted with two token counts: (i) 196 tokens, matching the RGB ViT-B/16 patch count, and (ii) 98 tokens, a 2$\times$ reduction. For GS, the token count corresponds to the number of Perceiver latent queries. For each GS configuration (splats $\in\{400,900,1600,3136\}$; tokens $\in\{98,196\}$), we apply both training stages described above. 

For RGB, we form the 98-token baseline by averaging each pair of patch embeddings in our pre-trained ViT-B/16 (Small) after the embedding and positional encoding, ensuring architectural parity with the GS setup. Although the fairest comparison would retrain an RGB ViT with 32×32 patches for 32 epochs, matching the GS model’s native 98-token regime, this is considerably more computationally costly. We therefore use this out-of-the-box reduction as a practical proxy; Appendix~D additionally reports finetuned 98-token RGB results, which further improve performance.

\subsection{Results} \label{results}


\begin{figure*}[htbp]
\begin{center}
\includegraphics[width=1.03\linewidth]{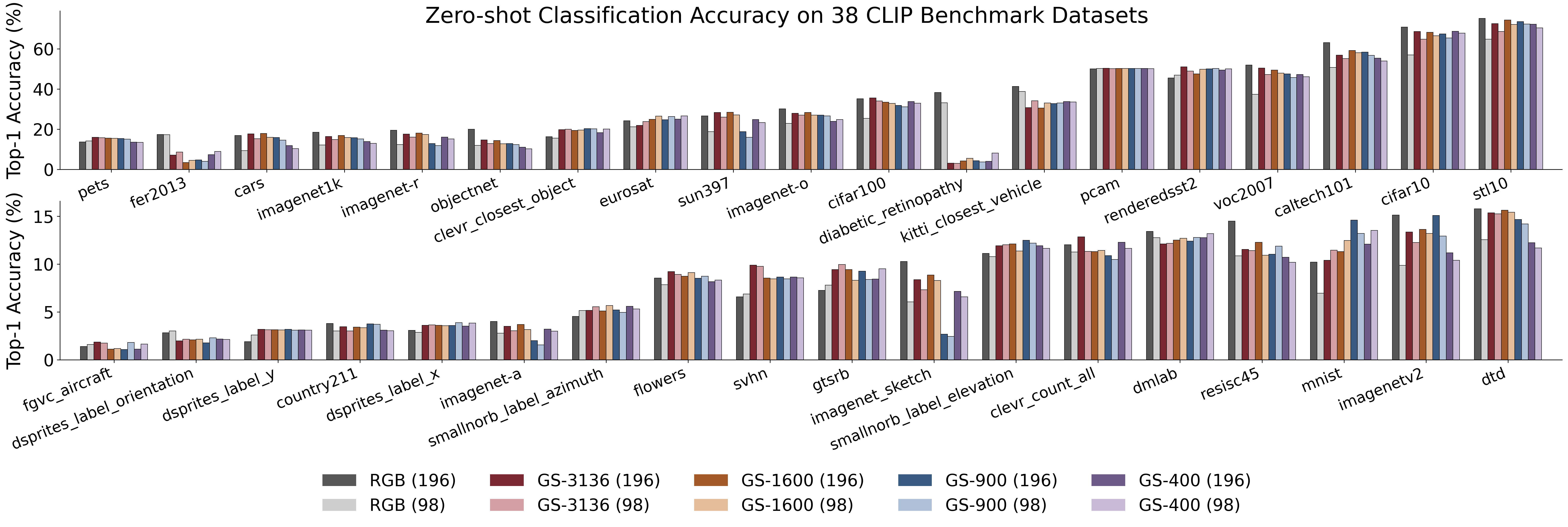}
\end{center}
\caption{Zero-shot classification accuracy on 38 datasets from the CLIP Benchmark for ViT-B-16 (Small) and multiple variants of GS vision encoders (number of gaussian points/img: 3136, 1600, 900, 400). Results are presented for 196 tokens (baseline) and 98 tokens.}
\label{accuracy}
\end{figure*}












\begin{table*}[htbp]
\centering
\renewcommand{\arraystretch}{1.05}
\setlength{\tabcolsep}{3.6pt}
\scriptsize

\begin{tabular}{lcccccccccc}
\toprule
\textbf{Vision Encoder} &
\multicolumn{2}{c}{\textbf{Representational Complexity}} &
\multicolumn{1}{c}{\textbf{Compression}} &
\multicolumn{2}{c}{\textbf{Data Load+Decode}} &
\multicolumn{2}{c}{\textbf{196 Token Accuracy}} &
\multicolumn{2}{c}{\textbf{98 Token Accuracy}} \\
\cmidrule(lr){2-3} \cmidrule(lr){4-4} \cmidrule(lr){5-6} \cmidrule(lr){7-8} \cmidrule(lr){9-10}
 & \textbf{Params/img} & \textbf{= Total} & \textbf{ratio} &
\textbf{Time (s)} & \textbf{Speedup} &
\textbf{Abs. avg.} & \textbf{Rel. to Base*} &
\textbf{Abs. avg.} & \textbf{Rel. to Base*} \\
\midrule

\textbf{Baseline} & & & & & & & & & \\

\quad RGB (224×224) &
224×224×3 &
150{,}528 &
1.00 &
0.094 &
1.00 &
22 &
\textbf{1.00} &
19 &
\textbf{0.87} \\

\midrule

\textbf{GS variants (Nb. points)} & & & & & & & & & \\

\quad GS (3136) &
3136×8 &
25{,}088 &
3.00 &
0.014 &
6.71 &
20 &
\textbf{0.98} &
20 &
\textbf{0.96} \\

\quad GS (1600) &
1600×8 &
12{,}800 &
5.88 &
0.007 &
13.43 &
20 &
\textbf{0.96} &
20 &
\textbf{0.95} \\

\quad GS (900) &
900×8 &
7{,}200 &
10.45 &
0.005 &
18.80 &
20 &
\textbf{0.92} &
19 &
\textbf{0.91} \\

\quad GS (400) &
400×8 &
3{,}200 &
23.52 &
0.003 &
31.33 &
19 &
\textbf{0.91} &
19 &
\textbf{0.92} \\

\bottomrule
\end{tabular}


\caption{Comparison between RGB and Gaussian Splat representations in terms of compression ratio, data loading and decoding speed, and average accuracy using 196 and 98 visual tokens. Abs. avg. and Rel. to Base* indicate the absolute mean accuracy of the encoder and the average relative accuracy respectively (explained in \ref{results}). Results in table form are provided in Appendix~D.}
\label{tab:gs_clip_compression}
\end{table*}

Fig.~\ref{accuracy} summarizes zero-shot accuracy across the 38 benchmarks, and Table~\ref{tab:gs_clip_compression} reports a comparative analysis of the compression, communication speedup, and relative accuracy. Compression ratios are computed relative to the uncompressed RGB input (224$\times$224$\times$3 bytes), following
\[
\textstyle
\text{Compression} 
= \frac{224\times224\times3\times1\text{B}}{N_{\text{GS}}\times 8\times 2\text{B}},
\]
where $N_{\text{GS}}$ denotes the number of Gaussian points and each splat stores 8 FP16 parameters (2 bytes each). This provides a conservative lower bound: prior work on Gaussian quantization (e.g., GaussianImage \cite{Zhang2024GaussianImage}) reports far more aggressive achievable compression. Data loading and decoding timings are measured on batches of 256 samples. All compression, throughput, and relative-accuracy metrics are reported with respect to our RGB ViT-B/16 (Small) baseline, which consumes 224$\times$224 pixel inputs and produces the standard 196 visual tokens. For each model variant, we report the mean zero-shot accuracy across the 38 datasets to assess generalization under distribution shift, as well as a \emph{relative accuracy} metric computed as the dataset-wise mean of 
${Acc}_{{model}}/{Acc}_{{baseline}}$. GS models achieve competitive alignment despite extreme compression. The 3136- and 1600-point GS models attain $96$–$98\%$ of the RGB baseline accuracy while reducing visual input size by $3\times$–$6\times$ and improving data loading speed by up to $13\times$. Even the 900- and 400-point variants preserve around $90$–$92\%$ of baseline accuracy at $10\times$–$23.5\times$ compression. Additionally, token reduction from 196→98 has only minor impact on GS models ($<2\%$ relative), suggesting that GS representations enjoy condensed semantic signal. Overall, 2DGS emerges as a compact, interpretable visual representation that is \emph{fluidly compatible} with pretrained RGB backbones and delivers strong alignment while substantially reducing storage, decoding cost, (and optionally token count). 



\section{Discussion}
\label{sec:conclusion}
Our study reveals both promise and clear limitations of 2DGS as a vision–language substrate. First, while Gaussian splats drastically reduce storage and data-loading time once computed, the fitting process itself remains non-trivial. Despite batched CUDA kernels and parallelization, our iterative optimizer still requires on the order of a few seconds per batch to produce high-quality splats at scale. Hardware acceleration has the potential to extend the scalability of 2DGS. A second limitation is modeling: training CLIP-style encoders from scratch from 2DGS inputs converges poorly compared to RGB, indicating that our current architectures and positional encodings are not yet optimally matched to splat geometry. Our two-stage recipe mitigates this by transferring from an RGB-pretrained ViT-B/16 backbone, and we show that, under this regime, GS often matches or even outperforms the RGB baseline on several individual datasets. However, this comes with two caveats: (i) GS performance is fundamentally upper-bounded by the quality of the underlying RGB teacher, and (ii) GS appears more sensitive to sharp distribution shifts, leading to slightly lower average accuracy across the full 38-dataset benchmark. This suggests that we have not yet discovered GS-native architectures with the same level of built-in inductive bias and robustness that modern ViTs enjoy, and that improving generalization under domain shift is a key open direction. Finally, our current alignment pipeline operates on non-pruned FP16 Gaussian parameters, leaving headroom for even stronger compression ratios.

\section{Conclusion}
This work demonstrates that 2D Gaussian Splatting can serve as an efficient and effective visual substrate for large-scale vision–language alignment. We introduced both algorithmic and systems optimizations to enable GS fitting and large-scale CLIP training, and proposed a two-stage alignment strategy that leverages the inductive strengths of RGB-pretrained ViTs. Across 38 zero-shot benchmarks, GS encoders retain $90$–$98\%$ of RGB accuracy while providing $3\times$–$23.5\times$ compression and large data-loading speedups, demonstrating that much of the semantic signal required for alignment can be retained in a compact, spatially adaptive representation, supporting the case for visual inputs that are better matched to both compute and communication constraints than dense pixels. More broadly, our results show that Gaussian splats are not merely a rendering primitive but a promising representational layer for multimodal learning. We view this work as a foundation for a new class of representation-first multimodal systems that begin with compact structure rather than compressing it after the fact.

\section*{Acknowledgements}
We thank the staff of the Stanford Marlowe computing cluster, in particular Kurt Stine and Craig Kapfer, for their continued support during our project. 
We are also grateful to NVIDIA solutions architects Zoe Ryan and Amanda Butler for their assistance with large-scale profiling and speedup. 
Finally, we thank Yuhui Zhang, Kedar Tatwawadi, Suresh Nambi, Ludwig Schmidt, Fernando Mujica, Mert Pelanci, and Marcel Rod for their insightful feedback and discussions.
{
    \small
    \bibliographystyle{ieeenat_fullname}
    \bibliography{main}
}


\onecolumn

\section*{Appendix}

The appendix provides an expanded set of experiments, analyses, and ablations that complement the main paper. To support full reproducibility, we will open-source our entire codebase, including 2DGS preprocessing, CUDA kernels, and CLIP training pipelines upon publication.


\tableofcontents
\vspace{1em}

\section{Additional Background}
\subsection{Contrastive Language-Image Pre-training}

Vision–language alignment is commonly achieved through Contrastive Language–Image Pre-Training (CLIP), which trains a dual-encoder architecture on image–caption pairs. A text encoder and a vision encoder are optimized jointly so that embeddings of aligned pairs are pulled closer, and mismatched pairs pushed apart under a contrastive loss \cite{Radford2021CLIP}. CLIP has become the standard paradigm, demonstrating strong performance across a wide range of downstream multimodal tasks. The CLIP framework spans a family of models from lightweight 25M-parameter encoders to larger 300M+ parameter variants such as ViT-L/14, all of which require hundreds of millions of training samples and significant compute to reach high-quality alignment.In this work, we retain the CLIP training recipe, but investigate whether 2D Gaussian Splatting can serve as an alternative visual substrate. Using the OpenCLIP implementation \cite{OpenCLIP2021}, we include comparisons of RGB- and 2DGS-based encoders under identical training conditions, isolating the effect of the representation itself. Our approach further reuses frozen RGB-pretrained backbones with lightweight splat-aware input modules, enabling efficient adaptation while substantially reducing trainable parameters.

\subsection{CLIP Benchmark.}
The \textit{CLIP Benchmark} is an evaluation suite for CLIP-like vision-language models, focusing on zero-shot performance across diverse tasks. In zero-shot evaluation, a pre-trained model is tested on new tasks without fine-tuning, using only natural language prompts for each target class. This provides a proxy measure of the encoder’s generalization ability and representation quality
dblp.org
dblp.org
. CLIP models \cite{Radford2021CLIP} have demonstrated strong zero-shot classification results on numerous image recognition datasets by simply using the class names or descriptions as text inputs, indicating the efficacy of their learned representations in transferring to unseen tasks.

The CLIP Benchmark \cite{clipbenchmark} encompasses a broad range of vision tasks to thoroughly assess such generalization. It includes standard zero-shot image classification datasets spanning various domains, as well as multi-label classification and image-text retrieval tasks. Notably, it incorporates all 19 tasks from the \textit{Visual Task Adaptation Benchmark (VTAB)} \cite{zhai2019}, a suite of classification tasks designed to evaluate general visual representations across heterogeneous domains. VTAB’s tasks are grouped into three categories, \textit{Natural}, \textit{Specialized}, and \textit{Structured},covering everything from everyday natural images to remote sensing and medical images, and even synthetic tasks that require counting objects or estimating distances. In addition to the VTAB tasks, the CLIP Benchmark evaluates models on even more complex zero-shot classification datasets. Table~\ref{tab:clip-tasks} summarizes the key datasets included, the type of task each represents, number of classes, and a brief description of what capability or scenario each dataset tests.

\begin{longtable}{p{3.2cm} p{3.8cm} p{1.2cm} p{8.0cm}}

\toprule
\textbf{Dataset} & \textbf{Task/Domain Type} & \textbf{\# Classes} & \textbf{Description of Task} \\
\midrule
\endfirsthead

\multicolumn{4}{c}%
{\tablename\ \thetable\ -- \textit{Continued from previous page}} \\
\toprule
\textbf{Dataset} & \textbf{Task/Domain Type} & \textbf{\# Classes} & \textbf{Description of Task} \\
\midrule
\endhead

\midrule
\multicolumn{4}{r}{\textit{Continued on next page}} \\
\endfoot

\bottomrule
\endlastfoot

ImageNet-1k & Object classification (natural) & 1000 & Standard ImageNet object recognition benchmark (ILSVRC-2012). \\ \midrule

ImageNet-v2 & Object classification (shifted) & 1000 & Re-collection of ImageNet validation set for distribution shift evaluation. \\ \midrule

ImageNet-R & Object classification (renditions) & 200 & Contains renditions of ImageNet categories in artistic or abstract styles (cartoons, paintings, sculptures). \\ \midrule

ImageNet-Sketch & Object classification (sketches) & 1000 & Sketch drawings of ImageNet classes, testing robustness to line-art style inputs. \\ \midrule

ObjectNet & Object classification (viewpoints) & 113 & Photos of objects from unusual viewpoints/backgrounds; tests robustness to pose and context changes. \\ \midrule

ImageNet-A & Object classification (adversarial) & 200 & "Naturally adversarial'' real-world images curated to fool standard ImageNet models (hard OOD test). \\ \midrule

CIFAR-10 & Object classification (low-res) & 10 & Tiny $32\times32$ natural images of 10 object classes (vehicles, animals). \\ \midrule

CIFAR-100 & Object classification (low-res) & 100 & Tiny $32\times32$ images across 100 fine-grained object categories. \\ \midrule

MNIST & Digit classification & 10 & Handwritten digit images (0--9) in grayscale. \\ \midrule

Oxford Flowers-102 & Fine-grained classification & 102 & Photographs of flowers; classify 102 flower species. \\ \midrule

Stanford Cars & Fine-grained classification & 196 & High-resolution photos of cars labeled by make, model, and year. \\ \midrule

SVHN & Digit classification (street images) & 10 & Street View House Numbers cropped into digits 0--9. \\ \midrule

FER-2013 & Facial emotion recognition & 7 & Low-resolution grayscale face images labeled with 7 expression categories. \\ \midrule

Rendered SST-2 & Text sentiment (OCR) & 2 & Images of text rendered from SST-2 movie reviews; classify sentiment. \\ \midrule

Oxford-IIIT Pets & Fine-grained classification & 37 & Photos of 37 cat and dog breeds. \\ \midrule

Caltech-101 & Object classification (varied) & 101 & Images of 101 diverse object categories (plus background). \\ \midrule

PASCAL VOC 2007 (Clf.) & Object presence (multi-label) & 20 & Detect presence/absence of 20 object categories in natural scenes. \\ \midrule

SUN397 & Scene classification & 397 & Scene recognition across 397 indoor/outdoor categories (park, office, bedroom, \dots). \\ \midrule

FGVC Aircraft & Fine-grained classification & 100 & Recognition of aircraft model variants across 100 categories. \\ \midrule

Country211 & Geographic location classification & 211 & Predict the country where the image was taken (211 possible labels). \\ \midrule

Describable Textures (DTD) & Texture classification & 47 & Classify images into 47 describable texture attributes (striped, dotted, grooved). \\ \midrule

GTSRB & Traffic sign recognition & 43 & Images of 43 German traffic sign classes captured from real-world road scenes. \\ \midrule

STL-10 & Object classification & 10 & Larger $96\times96$ version of CIFAR-like images with 10 object classes. \\ \midrule

Diabetic Retinopathy & Medical image classification & 5 & Classify retinal fundus images into 5 disease severity levels. \\ \midrule

EuroSAT & Satellite image classification & 10 & 10 land-use classes from Sentinel-2 satellite imagery (forest, farmland, river, \dots). \\ \midrule

RESISC45 & Aerial scene classification & 45 & Remote sensing scenes across 45 diverse aerial categories. \\ \midrule

PatchCamelyon (PCam) & Medical image classification & 2 & Microscopy patches labeled as tumor vs.\ normal. \\ \midrule

CLEVR Counts & Synthetic reasoning (counting) & 8 & Count objects in synthetic CLEVR 3D scenes (8 count categories). \\ \midrule

CLEVR Distances & Synthetic reasoning (spatial) & 6 & Predict relative distance of closest object in CLEVR scenes (6 bins). \\ \midrule

dSprites Orientation & Synthetic visual factor (orientation) & 40 & Classify rotation angle of a simple 2D shape. \\ \midrule

dSprites Position & Synthetic visual factor (position) & 32 & Classify object position on a grid (32 discrete locations). \\ \midrule

SmallNORB Elevation & Synthetic visual factor (3D pose) & 9 & Estimate camera elevation angle over 9 categories. \\ \midrule

SmallNORB Azimuth & Synthetic visual factor (3D pose) & 18 & Estimate camera azimuth angle over 18 viewpoints. \\ \midrule

DMLab & Synthetic visual reasoning (depth) & 6 & Classify 3D maze frames into 6 depth-related categories. \\ \midrule

KITTI Distances & Driving vision (depth estimation) & 4 & Classify distance to nearest vehicle into 4 discrete range bins. \\

\caption{Datasets covered in the CLIP zero-shot image classification benchmark \cite{cherti_2025_15403103} Each dataset’s task type, number of classes, and a brief description of what is evaluated are given.}
\label{tab:clip-tasks}

\end{longtable}

\vspace{2mm}

In summary, the CLIP Benchmark evaluates zero-shot transfer performance on a comprehensive collection of vision datasets. High accuracy across these diverse tasks (without task-specific training) indicates that a model has learned versatile and general visual features. Modern CLIP models (including large OpenCLIP variants trained on LAION-5B) indeed show strong performance on natural and specialized image classification tasks.


\section{Algorithmic Optimizations: Additional Details}

\subsection{Structured Initialization}

\begin{figure*}[h!]
\centering

\begin{tabular}{c c c c c c c}
    \begin{minipage}{0.17\linewidth}\centering
        {\small \textbf{Random Init}}\\[0.1cm]
        \includegraphics[width=\linewidth]{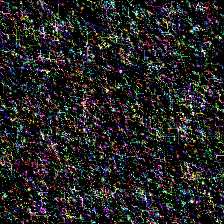}\\[0.1cm]
        {\small Iteration = 0}
    \end{minipage}
    &
    \begin{minipage}{0.17\linewidth}\centering
        {\small \textbf{Structured Init}}\\[0.1cm]
        \includegraphics[width=\linewidth]{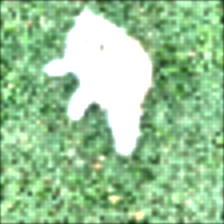}\\[0.1cm]
        {\small Iteration = 0}
    \end{minipage}
    &
    \hspace{0.03cm}
    &
    \begin{minipage}{0.17\linewidth}\centering
        {\small \textbf{Random Init}}\\[0.1cm]
        \includegraphics[width=\linewidth]{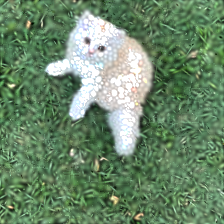}\\[0.1cm]
        {\small Iteration = 2000}
    \end{minipage}
    &
    \begin{minipage}{0.17\linewidth}\centering
        {\small \textbf{Structured Init}}\\[0.1cm]
        \includegraphics[width=\linewidth]{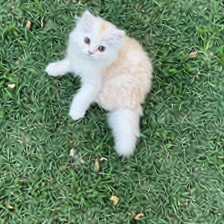}\\[0.1cm]
        {\small Iteration = 2000}
    \end{minipage}
    &
    \begin{minipage}{0.24\linewidth}\centering
        \includegraphics[width=\linewidth]{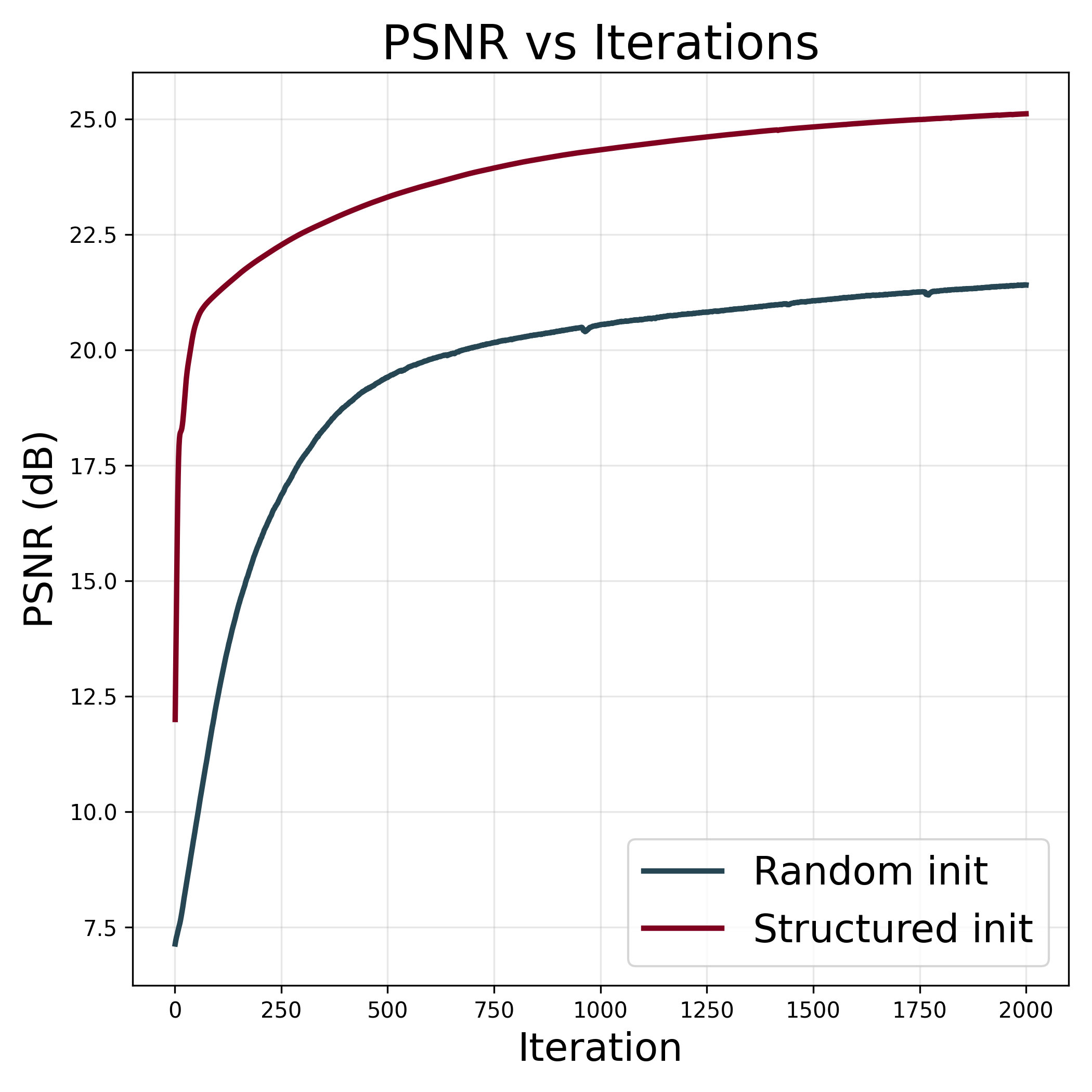}
        \\[-0.3em]
        {\small \textbf{PSNR vs. Iterations}}
    \end{minipage}
\end{tabular}

\caption{Structured vs. Random Initialization for 2DGS Fitting.
Left: qualitative comparison at iteration~0 and iteration~2000 under random and structured initialization (using 3136 Gaussian points).
Right: PSNR evolution during optimization. Structured initialization rapidly improves fidelity and consistently outperforms random initialization.}
\label{fig:structured_vs_random}
\end{figure*}

Figure \ref{fig:structured_vs_random} demonstrates the effectiveness of our structured initialization. Unlike random initialization, which begins from a noise-like configuration, our spatial layout-based initialization produces a meaningful coarse approximation at iteration~0. This strong prior accelerates optimization and yields consistently higher fidelity throughout training, as reflected in the PSNR curves in Fig. \ref{fig:structured_vs_random} (right). By iteration~2000, structured initialization produces sharper, more stable reconstructions under identical Gaussian budgets, thus unlocking higher asymptotic perceptual quality.

\subsection{Pruning}

\begin{figure*}[h!]
\centering

\begin{tabular}{c c c c}

    \begin{minipage}{0.5\linewidth}
        \centering
        \includegraphics[width=\linewidth]{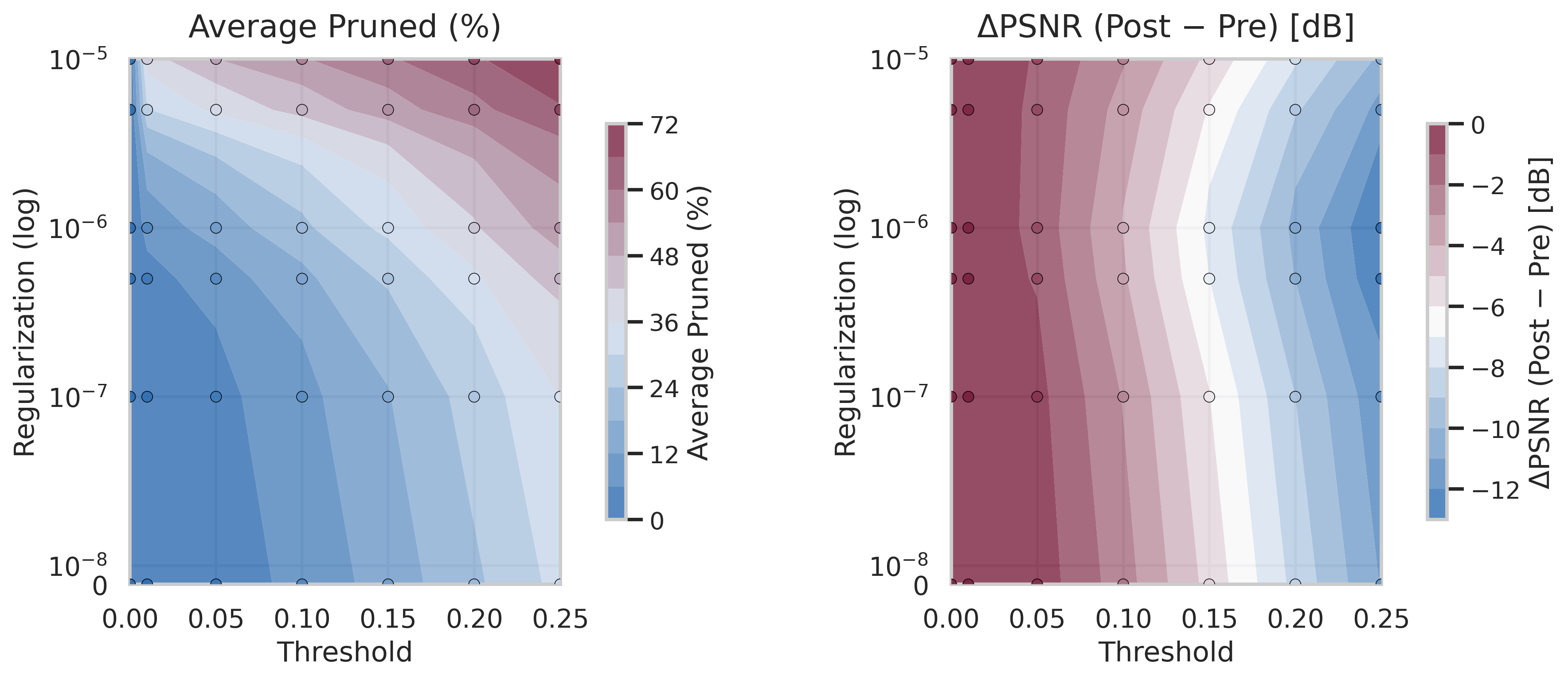}
    \end{minipage}
    &

    \begin{minipage}{0.2\linewidth}
        \centering
        \includegraphics[width=\linewidth]{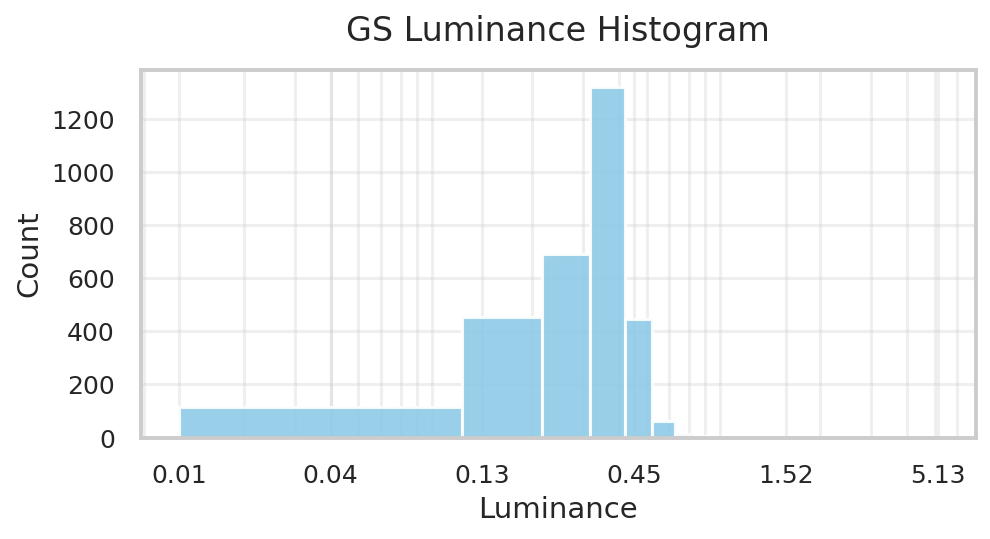}\\[0.6em]
        \includegraphics[width=\linewidth]{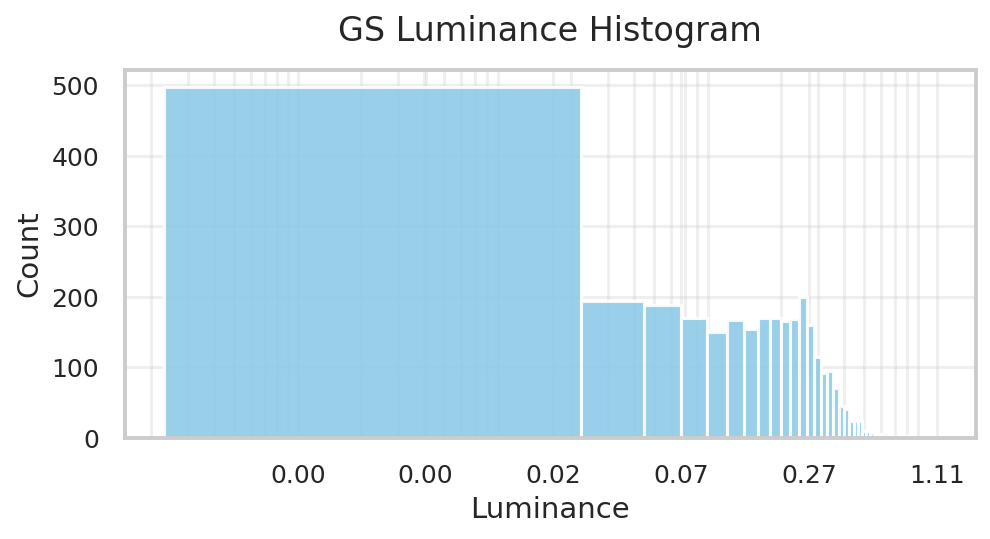}
    \end{minipage}
    &

    \begin{minipage}{0.1\linewidth}
        \centering
        \includegraphics[width=\linewidth]{new_figs/lam0_th0_perr.png}\\[0.6em]
        \includegraphics[width=\linewidth]{new_figs/lam1e-6_th0.05_prun23.72_psnr31.1.png}
    \end{minipage}
    &

    \begin{minipage}{0.1\linewidth}
        \centering
        \includegraphics[width=\linewidth]{new_figs/lam0_th0_psnr37.43_splat.png}\\[0.6em]
        \includegraphics[width=\linewidth]{new_figs/lam1e-6_th0.05_splat.png}
    \end{minipage}

\end{tabular}

\caption{Luminance behavior with color regularization and reconstruction quality across settings.
Left: contour plots visualizing pruning ratio and $\Delta$PSNR across threshold and regularization. Middle: luminance histograms before/after regularization. Right: Gaussian Splat and original RGB comparisons.}
\label{fig:gs_luminance_layout}
\end{figure*}

Figure~\ref{fig:gs_luminance_layout} summarizes our pruning study and provides additional detail beyond Sec.~4.4. 
To generate the contour maps (left), we sweep over luminance thresholds 
$\tau_{\mathrm{th}} \in \{0, 0.01, 0.05, 0.10, 0.15, 0.20, 0.25\}$ and color regularization weights 
$\lambda_{\mathrm{reg}} \in \{0, 10^{-7}, 5\times10^{-7}, 10^{-6}, 5\times10^{-6}, 10^{-5}\}$ 
for each Gaussian budget $\{400, 900, 1600, 3136\}$. 
For every $(\lambda_{\mathrm{reg}}, \tau_{\mathrm{th}})$ configuration, we fit the model for 2000 iterations on 
100 Mini-ImageNet samples (224$\times$224), record the resulting sparsity level, and compute 
$\Delta\mathrm{PSNR} = \mathrm{PSNR}_{\mathrm{post}} - \mathrm{PSNR}_{\mathrm{pre}}$ to quantify the fidelity impact of pruning. 
The contour maps therefore visualize the \emph{averaged} relationship between pruning aggressiveness and reconstruction stability.

As shown in Fig.~\ref{fig:gs_luminance_layout}, pruning increases smoothly with stronger regularization penalties and larger luminance thresholds, while $\Delta$PSNR remains small for moderate to large point budgets, which tend to tolerate more aggressive pruning. 
On the right, we present the luminance histogram, reconstructed image, and Gaussian splat visualization for a 3136-point GS fit (2000 iterations):
Top corresponds to $\lambda_{\mathrm{reg}}{=}0$, $\tau_{\mathrm{th}}{=}0$ (0\% pruned: PSNR = 37.43). 
Bottom corresponds to $\lambda_{\mathrm{reg}}{=}10^{-6}$, $\tau_{\mathrm{th}}{=}0.05$ (23.72\% pruned: PSNR = 31.1).


\section{CUDA Optimizations: Profiling Results}

To support large-scale 2DGS preprocessing, we extended the public \texttt{gsplat2d}
codebase (from \cite{Zhang2024GaussianImage} and further developed by \cite{Zhu2025LIG}) with full batch-parallel execution, enabling thousands of images to
be fitted concurrently on a single GPU. While our main paper reports
overall throughput improvements, here we include additional profiling details
on a H100 GPU using NVIDIA Nsight Systems to better illustrate kernel behavior.

We report profiling results of our revamped 2DGS CUDA pipeline under the configuration: 4000
Gaussian points and 2000 iterations on 224×224
mini-ImageNet images. Table~\ref{tab:nsys_cuda} reports the total fitting time
per batch, as well as Nsight Systems’ breakdown of GPU active time into compute
kernels vs.\ memory operations.

\begin{table}[h]
\centering
\begin{tabular}{lcccc}
\toprule
\textbf{Batch size} &
\textbf{Total fit time} &
\textbf{Kernel} &
\textbf{Memory} &
\textbf{Speedup} \\
\midrule
1     & 7 s     & 91.9\% & 8.1\%  & 1 \\
128   & 13 s    & 97.9\% & 2.1\%  & 1.86 \\
256   & 22 s    & 92.7\% & 7.3\%  & 3.14 \\
512   & 39 s    & 93.2\% & 6.8\%  & 5.57 \\
1024  & 1m 16s  & 89.9\% & 10.1\% & 10.86 \\
2048  & 2m 23s  & 94.6\% & 5.4\%  & 20.43 \\
4096  & 5m 27s  & 81.2\% & 18.8\% & 46.71 \\
\bottomrule
\end{tabular}
\caption{\textbf{Nsight Systems profiling of our batch-parallel 2DGS CUDA kernels.}
Reported numbers correspond to 4000 points and 2000 iterations per
image. Kernel/Memory percentages reflect the fraction of \emph{active} GPU time.}
\label{tab:nsys_cuda}
\end{table}

Not surprisingly (by Amdahl's law), the wall-clock fitting time grows
sublinearly with batch size, helping us identify cost-effective operating points
when scaling to the full \textbf{12.8M-image Datacomp dataset}. When fitting at scale, for each point
budget (400, 900, 1600, 3136, and 4000), we choose a distinct batch
size that maximizes throughput; the final batch sizes are reported in
the Data Pre-processing section.

\section{Vision-Language Alignment at Scale from 2D Gaussian Splat Representations: Additional Details}

\subsection{Training Data Pre-processing through 2D Gaussian Splatting}

To train our models efficiently, we pre-fit the entire 12.8M-image DataComp dataset offline using 2D Gaussian Splatting. Although per-image fitting is significantly faster with 2DGS compared to classical INRs, it remains too costly to perform online. Offline fitting allows us to amortize this one-time cost across all downstream experiments and iterate rapidly on model design.

For each configuration in Table~\ref{tab:gs_stats}, the ``Config.'' entry corresponds to the number of Gaussian points per image. We run each fit for \textbf{2000 iterations}, a conservative setting chosen, because PSNR consistently plateaus before that point. Batch sizes were selected via profiling to maximize GPU throughput for each configuration. Total GPU hours (total time to fit the entire 12.8M dataset) were obtained by dividing the full dataset size by the measured wall-clock time per batch.

We additionally report dataset statistics for each configuration. Specifically, we collect the empirical means and standard deviations of the covariance components $(\mathrm{cov}_{xx}, \mathrm{cov}_{xy}, \mathrm{cov}_{yy})$ and RGB channels from a 1M-image subset. These statistics support the normalization strategies explored later in the appendix. Figure~\ref{fig:gs_quality} visualizes the perceptual reconstructions across configurations, highlighting that, even when perceptual fidelity degrades at lower point counts (eg. 400 Gaussian points), the semantic signal may remain largely intact for many downstream tasks.

\begin{table}[htbp]
\centering
\small
\renewcommand{\arraystretch}{1.2}
\setlength{\tabcolsep}{4pt}

\begin{tabular}{p{1.7cm} p{1.4cm} p{2.0cm} cccc}
\toprule
\textbf{Config.} &
\textbf{Bsz} &
\textbf{Total GPU hrs} &
\multicolumn{4}{c}{\textbf{Data Statistics}} \\
\cmidrule(lr){4-7}
& & &
\textbf{cov mean} &
\textbf{cov std} &
\textbf{rgb mean} &
\textbf{rgb std} \\
\midrule

400 & 4096 & 25.6 &
$[\,3.40,\;-0.01,\;3.38\,]$ &
$[\,0.37,\;1.65,\;0.39\,]$ &
$[\,0.77,\;0.74,\;0.73\,]$ &
$[\,1.52,\;1.48,\;1.50\,]$ \\

900 & 2048 & 42.8 &
$[\,2.60,\;-4.45\mathrm{e}{-3},\;2.58\,]$ &
$[\,0.43,\;1.19,\;0.44\,]$ &
$[\,0.66,\;0.63,\;0.62\,]$ &
$[\,0.97,\;0.96,\;0.96\,]$ \\

1600 & 2048 & 35.9 &
$[\,2.05,\;-2.53\mathrm{e}{-3},\;2.07\,]$ &
$[\,0.32,\;0.77,\;0.30\,]$ &
$[\,0.58,\;0.56,\;0.55\,]$ &
$[\,0.59,\;0.58,\;0.59\,]$ \\

3136 & 1024 & 53.1 &
$[\,1.63,\;-1.48\mathrm{e}{-3},\;1.65\,]$ &
$[\,0.30,\;0.58,\;0.29\,]$ &
$[\,0.49,\;0.48,\;0.46\,]$ &
$[\,0.46,\;0.45,\;0.46\,]$ \\

\bottomrule
\end{tabular}
\caption{\textbf{Gaussian Splat fitting configurations.} 
Each configuration specifies the number of Gaussian points per image. 
Batch sizes were selected using CUDA profiling for maximal throughput. 
Total GPU hours are estimated by dividing the 12.8M-image dataset size by the measured time per batch. 
We report dataset-level statistics (mean and std) for covariance components $(\mathrm{cov}_{xx},\mathrm{cov}_{xy},\mathrm{cov}_{yy})$ and RGB channels, computed over 1M fitted images.}
\label{tab:gs_stats}
\end{table}

\begin{figure*}[h!]
\centering

\begin{subfigure}{0.19\textwidth}
    \centering
    \includegraphics[width=\linewidth]{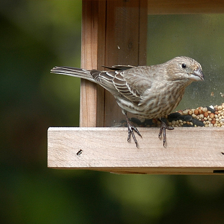}
    \caption{Original}
\end{subfigure}
\begin{subfigure}{0.19\textwidth}
    \centering
    \includegraphics[width=\linewidth]{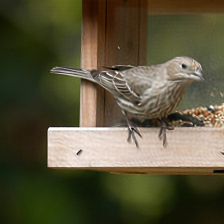}
    \caption{GS: 3136 pts}
\end{subfigure}
\begin{subfigure}{0.19\textwidth}
    \centering
    \includegraphics[width=\linewidth]{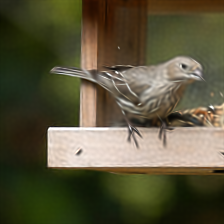}
    \caption{GS: 1600 pts}
\end{subfigure}
\begin{subfigure}{0.19\textwidth}
    \centering
    \includegraphics[width=\linewidth]{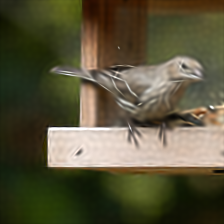}
    \caption{GS: 900 pts}
\end{subfigure}
\begin{subfigure}{0.19\textwidth}
    \centering
    \includegraphics[width=\linewidth]{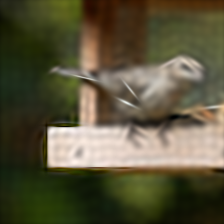}
    \caption{GS: 400 pts}
\end{subfigure}
\caption{\textbf{Perceptual reconstructions across Gaussian point budgets.}
Although perceptual fidelity decreases at lower point counts (higher compression ratios), the underlying semantic content often remains largely preserved, enabling effective downstream learning.}
\label{fig:gs_quality}
\end{figure*}

\subsection{Model Architecture Details}
Our GaussianSplatEncoder maps a set of $N$ Gaussian splat primitives
$(x,y,\mathrm{cov}_{xx},\mathrm{cov}_{xy},\mathrm{cov}_{yy},R,G,B)$
to a CLIP-compatible embedding. The number of Gaussians $N$ (e.g., 196, 400, 900, 1600, 3136) and the number of latent tokens $M$ (e.g., 196 or 98) are fully configurable.
The GaussianSplatEncoder is composed of a GSStem (Perceiver Resampler-based architecture), followed by a transformer backbone and projection. Each 8-D Gaussian point is normalized (XY scaling, signed--log covariance) and Fourier-encoded with $6$ frequencies (Fourier dim $=4\!\times\!6$). A linear layer projects the input to a \textbf{128-d} Perceiver space. The stem uses learnable latent queries of shape $M\times128$ and applies a stack of \textbf{4 cross-attention layers}, each with 4 heads. The output is projected to the CLIP width of 512. The resampled $M$ tokens are pre-pended with a learned CLS token and fed to a transformer backbone with width = 512, layers $L$ (default 12), heads $H$ (default 8),
MLP ratio = 4, RMSNorm pre-normalization. The CLS output is layer-normalized and projected to the final 512-d embedding.


\begin{algorithm}[h]
\caption{GaussianSplatEncoder (forward)}
\begin{algorithmic}[1]
\Require $X \in \mathbb{R}^{B\times N\times 8}$, number of latents $M$
\State $X \gets$ normalize\_xy\_cov\_rgb$(X)$
\If{Fourier enabled}
    \State $X \gets$ fourier\_encode$(X)$
\EndIf
\State $X \gets \mathrm{Linear}_{8+\mathrm{fourier}\_\mathrm{dim} \rightarrow 128}(X)$
\State $\ell \gets$ learnable\_latents $(M\times128)$, broadcast to batch
\For{$t=1\ldots4$}
    \State $\ell \gets \mathrm{CrossAttn}(\ell, X)$
\EndFor
\State $T \gets \mathrm{Linear}_{128\rightarrow512}(\ell)$
\State prepend CLS token: $Z = [\mathrm{CLS};\, T]$
\State $Z \gets \mathrm{RMSNorm}(Z)$
\State $Z \gets \mathrm{Transformer}_{L,H,\mathrm{MLP}=4}(Z)$
\State $\mathbf{h} \gets \mathrm{LN}(Z_{\mathrm{CLS}})$
\State \Return $\mathrm{Linear}_{512\rightarrow512}(\mathbf{h})$
\end{algorithmic}
\end{algorithm}

\subsection{Zero-Shot Performance Results}

\begin{figure*}[h!]
    \centering
    \subfloat[CKA Mean]{
        \includegraphics[width=0.32\linewidth]{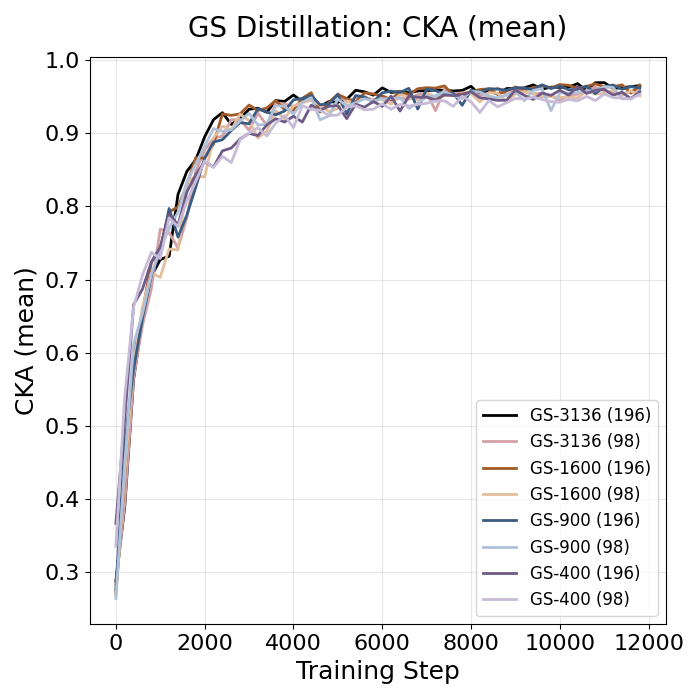}
    }
    \hfill
    \subfloat[Train Cosine]{
        \includegraphics[width=0.32\linewidth]{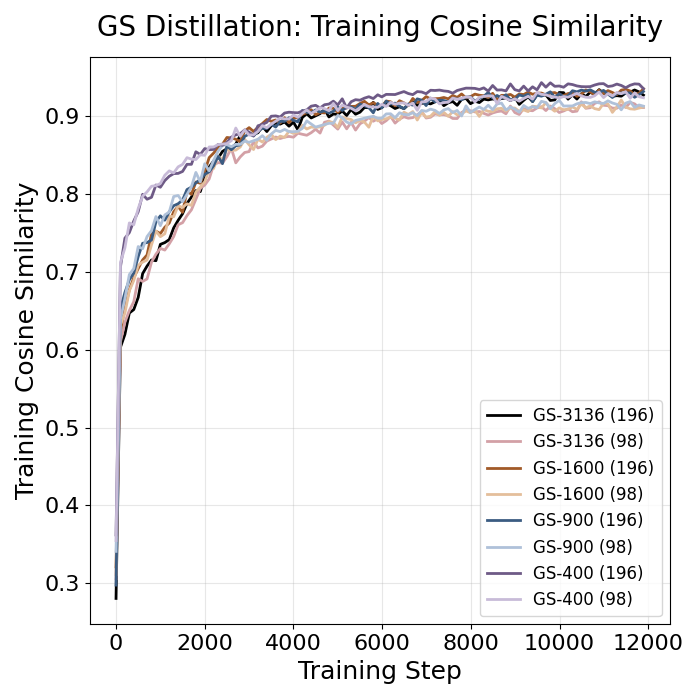}
    }
    \hfill
    \subfloat[Val Cosine]{
        \includegraphics[width=0.32\linewidth]{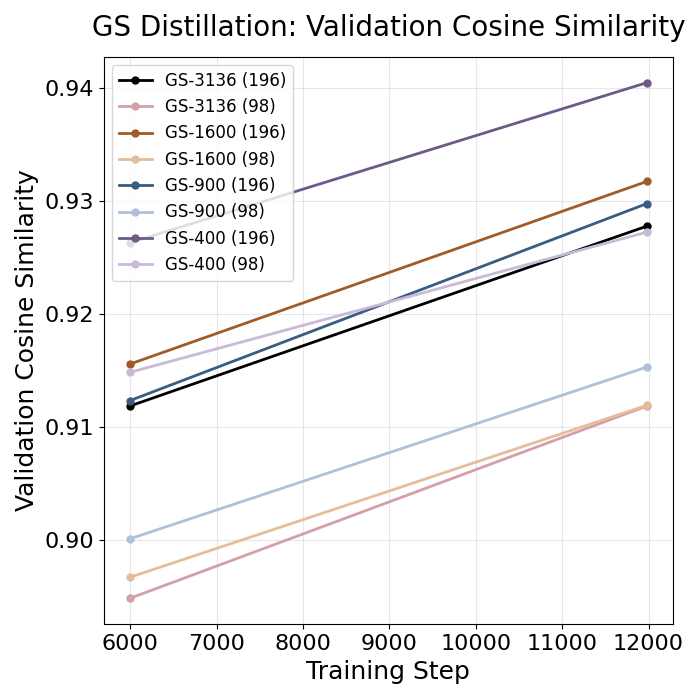}
    }
    \caption{Diagnostic distillation metrics illustrating representation alignment during CLIP training.}
    \label{fig:distill_diagnostics}
\end{figure*}

\begin{figure*}[h!]
    \centering
    \subfloat[Zero-shot Top-1 after 1 Epoch]{
        \includegraphics[width=0.48\linewidth]{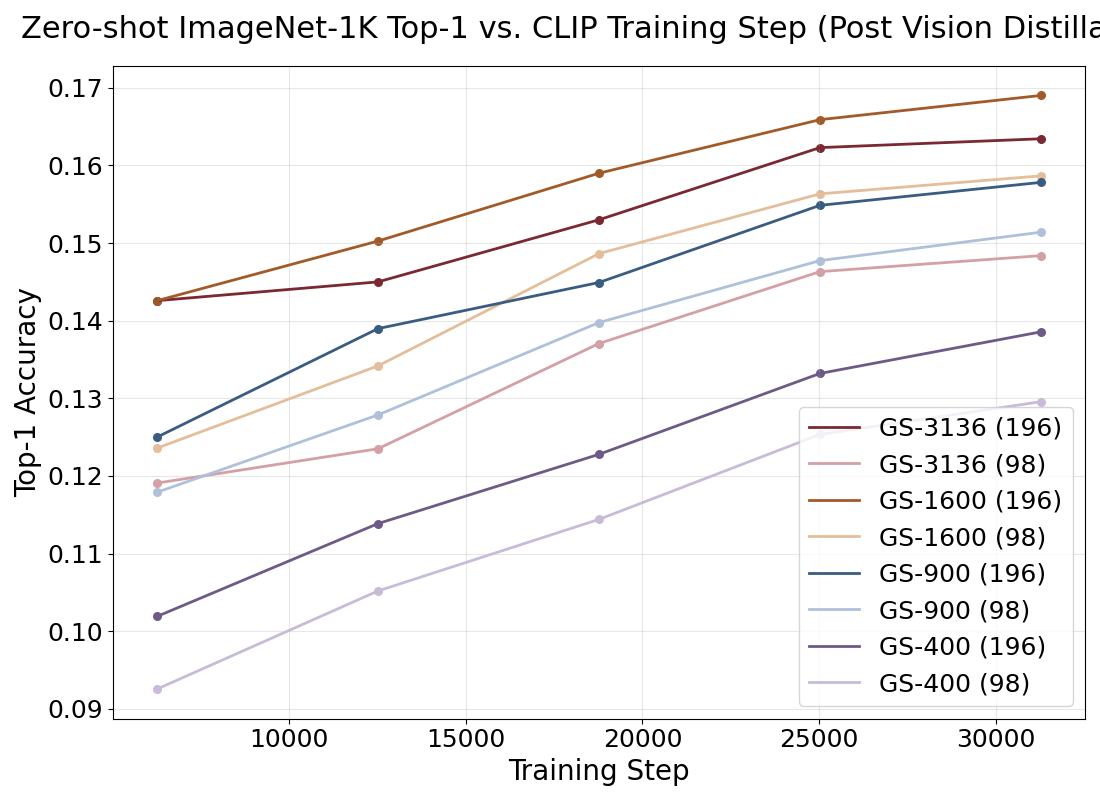}
    }
    \hfill
    \subfloat[Training Loss]{
        \includegraphics[width=0.48\linewidth]{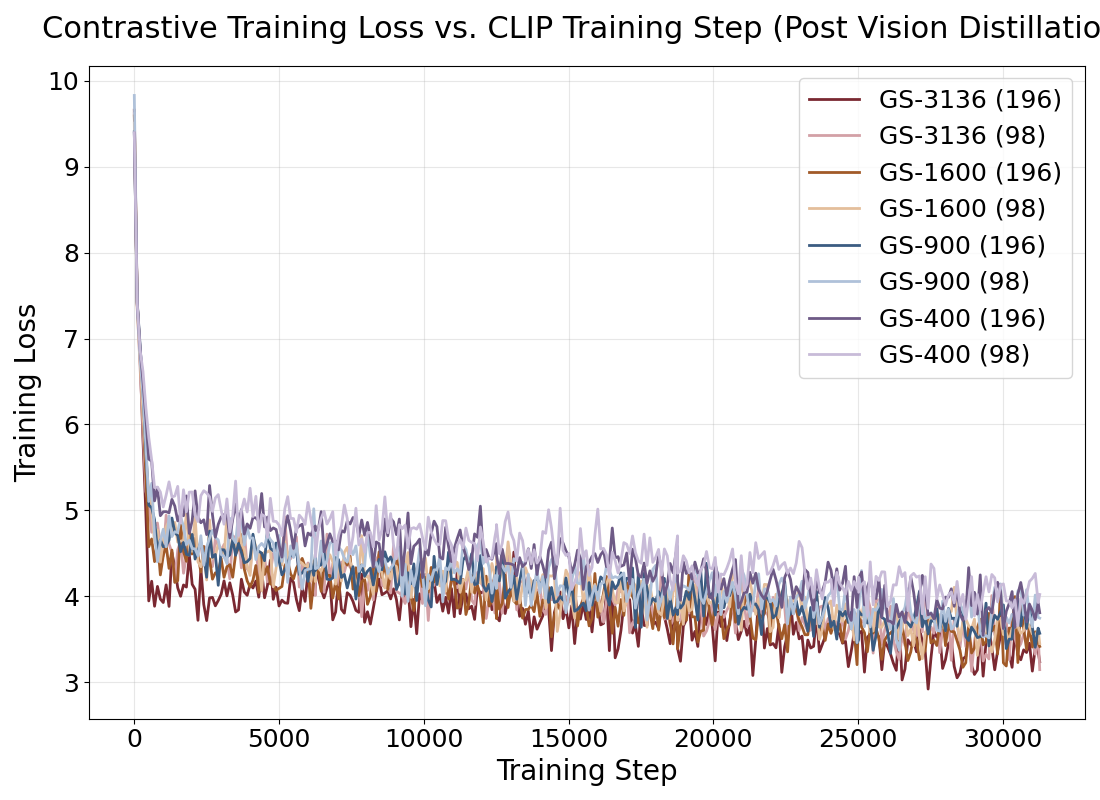}
    }
    \caption{Training loss and top-1 accuracy vs. training steps for our GS encoders.}
    \label{train_steps}
\end{figure*}

In the main paper, we presented our zero-shot accuracies in bar-plot form; for completeness, the full table of results is provided in Table~\ref{tab:acc} (breakdown of relative accuracies provided in Table~\ref{tab:rel}). We also include the distillation stage metrics in Fig. \ref{fig:distill_diagnostics} as well as the training loss and top-1 accuracy vs. training steps in \ref{train_steps}. We recall that RGB ViT-B/16 (196-tokens) is used as the baseline.

Across the 38 benchmark datasets, GS encoders exhibit competitive and often superior performance to the RGB ViT baseline. Notably, 19 datasets are best solved by one of the GS variants, demonstrating that the 2DGS representation preserves strong semantic information despite its aggressive compression. Even the 400-point configuration retains a surprising amount of semantic signal. 

We also observe two datasets where GS underperforms more substantially. These cases appear tied to stronger distribution shifts, where the ViT RGB model exhibits more robust generalization and higher absolute accuracy. As this work represents the first systematic exploration of 2D Gaussian Splatting for vision-language alignment, we expect that more mature GS-native architectures will further improve robustness under distribution shift, narrowing the performance gap with RGB-based models.

\begin{table*}[htbp]
\centering

\begin{tabular}{l
                cc
                cc
                cc
                cc
                cc}
\toprule
\textbf{Dataset} & 
\multicolumn{2}{c}{\textbf{RGB ViT-B/16 (Small)}} &
\multicolumn{2}{c}{\textbf{GS 3136}} &
\multicolumn{2}{c}{\textbf{GS 1600}} &
\multicolumn{2}{c}{\textbf{GS 900}} &
\multicolumn{2}{c}{\textbf{GS 400}} \\
\cmidrule(lr){2-3}
\cmidrule(lr){4-5}
\cmidrule(lr){6-7}
\cmidrule(lr){8-9}
\cmidrule(lr){10-11}
& 196 & 98 & 196 & 98 & 196 & 98 & 196 & 98 & 196 & 98 \\
\midrule

cars & 16.88 & 9.35 & \underline{17.73} & 15.33 & \textbf{17.95} & 15.96 & 15.87 & 14.60 & 11.85 & 10.36 \\
country211 & \textbf{3.79} & 3.01 & 3.45 & 3.01 & 3.42 & 3.34 & \underline{3.76} & 3.71 & 3.10 & 3.03 \\
fer2013 & \textbf{17.41} & \underline{17.34} & 7.19 & 8.62 & 3.46 & 4.49 & 4.79 & 4.08 & 7.33 & 8.99 \\
fgvc\_aircraft & 1.38 & 1.59 & \textbf{1.86} & 1.74 & 1.11 & 1.17 & 1.07 & \underline{1.81} & 1.11 & 1.65 \\
gtsrb & 7.25 & 7.81 & 9.44 & \textbf{9.96} & 9.44 & 8.31 & 9.27 & 8.41 & 8.43 & \underline{9.52} \\
imagenet-a & \textbf{4.00} & 2.79 & 3.51 & 3.04 & \underline{3.69} & 3.16 & 2.00 & 1.56 & 3.21 & 2.99 \\
imagenet-o & \textbf{30.20} & 22.95 & 27.90 & 27.00 & \underline{28.40} & 27.00 & 27.00 & 26.60 & 23.90 & 24.80 \\
imagenet-r & \textbf{19.46} & 12.40 & 17.58 & 16.07 & \underline{18.10} & 17.44 & 12.84 & 11.84 & 16.06 & 15.21 \\
imagenet1k & \textbf{18.51} & 12.14 & 16.40 & 14.89 & \underline{16.87} & 15.87 & 15.77 & 15.16 & 13.83 & 12.94 \\
imagenet\_sketch & \textbf{10.27} & 6.06 & 8.39 & 7.33 & \underline{8.87} & 8.29 & 2.69 & 2.44 & 7.14 & 6.59 \\
imagenetv2 & \textbf{15.14} & 9.88 & 13.37 & 12.27 & 13.63 & 13.21 & \underline{15.09} & 12.94 & 11.18 & 10.40 \\
mnist & 10.22 & 6.97 & 10.41 & 11.47 & 11.32 & 12.48 & \textbf{14.60} & 13.21 & 12.09 & \underline{13.53} \\
objectnet & \textbf{19.99} & 11.96 & \underline{14.64} & 12.92 & 14.41 & 12.75 & 12.86 & 12.36 & 11.11 & 10.22 \\
renderedsst2 & 45.52 & 46.95 & \textbf{51.07} & 48.98 & 47.56 & 49.86 & 50.08 & \underline{50.25} & 49.48 & 50.03 \\
stl10 & \textbf{75.17} & 64.79 & 72.56 & 68.66 & \underline{74.35} & 72.21 & 73.58 & 72.41 & 72.24 & 70.43 \\
sun397 & 26.66 & 18.83 & \underline{28.40} & 26.05 & \textbf{28.46} & 27.17 & 18.85 & 16.04 & 24.83 & 23.31 \\
voc2007 & \textbf{52.01} & 37.42 & \underline{50.49} & 47.27 & 49.49 & 47.98 & 47.55 & 45.78 & 47.27 & 46.12 \\

caltech101 & \textbf{63.16} & 50.71 & 56.84 & 55.20 & \underline{59.19} & 58.13 & 58.40 & 56.79 & 55.42 & 53.96 \\
cifar10 & \textbf{70.89} & 57.04 & 68.68 & 64.82 & 68.26 & 66.50 & 67.48 & 65.48 & \underline{68.78} & 67.87 \\
cifar100 & 35.15 & 25.39 & \textbf{35.62} & 34.11 & 33.47 & 32.91 & 31.91 & 31.13 & \underline{33.79} & 32.94 \\
clevr\_closest\_object\_distance & 16.29 & 15.58 & 19.83 & 19.97 & 19.42 & 19.63 & \textbf{20.31} & \underline{20.21} & 18.31 & 20.07 \\
clevr\_count\_all & 12.03 & 11.27 & \textbf{12.85} & 11.33 & 11.32 & 11.45 & 10.89 & 10.50 & \underline{12.29} & 11.65 \\
diabetic\_retinopathy & \textbf{38.33} & \underline{33.21} & 3.09 & 2.99 & 4.28 & 5.54 & 4.35 & 3.76 & 4.05 & 8.14 \\
dmlab & \textbf{13.42} & 12.77 & 12.12 & 12.19 & 12.52 & 12.70 & 12.40 & 12.78 & 12.77 & \underline{13.18} \\
dsprites\_label\_orientation & \underline{2.81} & \textbf{3.01} & 1.98 & 2.15 & 2.08 & 2.15 & 1.76 & 2.29 & 2.17 & 2.13 \\
dsprites\_label\_x\_position & 3.08 & 2.86 & 3.61 & 3.65 & 3.60 & 3.55 & 3.59 & \textbf{3.88} & 3.51 & \underline{3.85} \\
dsprites\_label\_y\_position & 1.88 & 2.60 & \textbf{3.18} & 3.14 & 3.15 & 3.12 & \underline{3.17} & 3.10 & 3.11 & 3.09 \\
dtd & \textbf{15.80} & 12.55 & 15.37 & 15.27 & \underline{15.64} & 15.43 & 14.68 & 14.20 & 12.23 & 11.70 \\
eurosat & 24.28 & 21.26 & 21.94 & 23.80 & 24.94 & \underline{26.59} & 24.72 & 26.35 & 25.05 & \textbf{26.61} \\
flowers & 8.55 & 7.84 & \textbf{9.22} & 8.93 & 8.73 & \underline{9.11} & 8.52 & 8.73 & 8.16 & 8.34 \\
kitti\_closest\_vehicle\_distance & \textbf{41.35} & \underline{38.82} & 30.80 & 34.18 & 30.52 & 33.05 & 32.77 & 33.05 & 33.76 & 33.61 \\
pcam & 50.10 & \underline{50.30} & \textbf{50.34} & 50.21 & 50.30 & 50.25 & 50.25 & 50.23 & 50.21 & 50.13 \\
pets & 13.65 & 14.17 & \textbf{15.94} & \underline{15.81} & 15.62 & 15.51 & 15.40 & 15.13 & 13.57 & 13.46 \\
resisc45 & \textbf{14.51} & 10.87 & 11.54 & 11.43 & \underline{12.29} & 10.94 & 11.04 & 11.89 & 10.73 & 10.19 \\
smallnorb\_label\_azimuth & 4.53 & 5.14 & 5.18 & 5.56 & 5.10 & \textbf{5.67} & 5.20 & 4.96 & \underline{5.59} & 5.31 \\
smallnorb\_label\_elevation & 11.13 & 10.78 & 11.93 & 12.03 & 12.12 & 11.38 & \textbf{12.50} & \underline{12.21} & 11.93 & 11.65 \\
svhn & 6.58 & 6.87 & \textbf{9.91} & \underline{9.76} & 8.54 & 8.45 & 8.65 & 8.45 & 8.65 & 8.57 \\

\midrule
\textbf{Absolute Average Accuracy} &
\textbf{22.20} & \textbf{18.52} &
\textbf{20.39} & \textbf{19.76} &
\textbf{20.31} & \textbf{20.07} &
\textbf{19.61} & \textbf{19.14} &
\textbf{19.41} & \textbf{19.37} \\
\bottomrule
\end{tabular}

\caption{Zero-shot classification accuracy across datasets. For each dataset, the best score (across all models and token counts) is shown in bold, and the second-best is underlined.}
\label{tab:acc}
\end{table*}

\begin{table*}[htbp]
\centering

\begin{tabular}{l
                cc
                cc
                cc
                cc
                cc}
\toprule
\textbf{Dataset} & 
\multicolumn{2}{c}{\textbf{RGB ViT-B/16 (Small)}} &
\multicolumn{2}{c}{\textbf{GS 3136}} &
\multicolumn{2}{c}{\textbf{GS 1600}} &
\multicolumn{2}{c}{\textbf{GS 900}} &
\multicolumn{2}{c}{\textbf{GS 400}} \\
\cmidrule(lr){2-3}
\cmidrule(lr){4-5}
\cmidrule(lr){6-7}
\cmidrule(lr){8-9}
\cmidrule(lr){10-11}
& 196 & 98 & 196 & 98 & 196 & 98 & 196 & 98 & 196 & 98 \\
\midrule

cars & 1.00 & 0.55 & \underline{1.05} & 0.91 & \textbf{1.06} & 0.95 & 0.94 & 0.86 & 0.70 & 0.61 \\

country211 & \textbf{1.00} & 0.79 & 0.91 & 0.79 & 0.90 & 0.88 & \underline{0.99} & 0.98 & 0.82 & 0.80 \\

fer2013 & \textbf{1.00} & \underline{1.00} & 0.41 & 0.50 & 0.20 & 0.26 & 0.28 & 0.23 & 0.42 & 0.52 \\

fgvc\_aircraft & 1.00 & \underline{1.15} & \textbf{1.35} & 1.26 & 0.80 & 0.85 & 0.78 & 1.31 & 0.80 & 1.20 \\

gtsrb & 1.00 & 1.08 & \underline{1.30} & \textbf{1.37} & \underline{1.30} & 1.15 & 1.28 & 1.16 & 1.16 & 1.31 \\

imagenet-a & 1.00 & 0.70 & 0.88 & 0.76 & \underline{0.92} & 0.79 & 0.50 & 0.39 & 0.80 & 0.75 \\

imagenet-o & 1.00 & 0.76 & \underline{0.92} & 0.89 & \textbf{0.94} & 0.89 & 0.89 & 0.88 & 0.79 & 0.82 \\

imagenet-r & 1.00 & 0.64 & 0.90 & 0.83 & \underline{0.93} & 0.90 & 0.66 & 0.61 & 0.83 & 0.78 \\

imagenet1k & 1.00 & 0.66 & 0.89 & 0.80 & \underline{0.91} & 0.86 & 0.85 & 0.82 & 0.75 & 0.70 \\

imagenet\_sketch & 1.00 & 0.59 & 0.82 & 0.71 & \underline{0.86} & 0.81 & 0.26 & 0.24 & 0.70 & 0.64 \\

imagenetv2 & \textbf{1.00} & 0.65 & 0.88 & 0.81 & 0.90 & 0.87 & \underline{1.00} & 0.85 & 0.74 & 0.69 \\

mnist & 1.00 & 0.68 & 1.02 & \underline{1.12} & 1.11 & 1.22 & \textbf{1.43} & 1.29 & 1.18 & 1.32 \\

objectnet & \textbf{1.00} & 0.60 & 0.73 & 0.65 & 0.72 & 0.64 & 0.64 & 0.62 & 0.56 & \underline{0.51} \\

renderedsst2 & 1.00 & 1.03 & \textbf{1.12} & \underline{1.08} & 1.04 & 1.10 & 1.10 & 1.10 & 1.09 & 1.10 \\

stl10 & 1.00 & 0.86 & 0.97 & 0.91 & \underline{0.99} & 0.96 & 0.98 & 0.96 & 0.96 & \textbf{0.94} \\

sun397 & 1.00 & 0.71 & \underline{1.07} & 0.98 & \textbf{1.07} & 1.02 & 0.71 & 0.60 & 0.93 & 0.87 \\

voc2007 & \textbf{1.00} & 0.72 & 0.97 & 0.91 & \underline{0.95} & 0.92 & 0.91 & 0.88 & 0.91 & 0.89 \\

caltech101 & \textbf{1.00} & 0.80 & 0.90 & 0.87 & \underline{0.94} & 0.92 & 0.92 & 0.90 & 0.88 & 0.85 \\

cifar10 & 1.00 & 0.80 & \underline{0.97} & 0.91 & 0.96 & 0.94 & 0.95 & 0.92 & \textbf{0.97} & 0.96 \\

cifar100 & 1.00 & 0.72 & \underline{1.01} & 0.97 & 0.95 & 0.94 & 0.91 & 0.89 & 0.96 & 0.94 \\

clevr\_closest\_object\_distance & 1.00 & 0.96 & 1.22 & \underline{1.23} & 1.19 & 1.21 & \textbf{1.25} & 1.24 & 1.12 & 1.23 \\

clevr\_count\_all & 1.00 & 0.94 & \textbf{1.07} & 0.94 & 0.94 & 0.95 & 0.91 & 0.87 & \underline{1.02} & 0.97 \\

diabetic\_retinopathy & \textbf{1.00} & \underline{0.87} & 0.08 & 0.08 & 0.11 & 0.14 & 0.11 & 0.10 & 0.11 & 0.21 \\

dmlab & \textbf{1.00} & 0.95 & 0.90 & 0.91 & \underline{0.93} & 0.95 & 0.92 & 0.95 & 0.95 & 0.98 \\

dsprites\_label\_orientation & \textbf{1.00} & \underline{1.07} & 0.70 & 0.77 & 0.74 & 0.77 & 0.63 & 0.81 & 0.77 & 0.76 \\

dsprites\_label\_x\_position & 1.00 & 0.93 & \underline{1.17} & 1.19 & 1.17 & 1.15 & 1.17 & \textbf{1.26} & 1.14 & 1.25 \\

dsprites\_label\_y\_position & 1.00 & \underline{1.38} & 1.69 & 1.67 & \textbf{1.68} & 1.66 & 1.69 & 1.65 & 1.65 & 1.64 \\

dtd & \textbf{1.00} & 0.79 & 0.97 & 0.97 & \underline{0.99} & 0.98 & 0.93 & 0.90 & 0.77 & 0.74 \\

eurosat & 1.00 & 0.88 & 0.90 & 0.98 & 1.03 & \underline{1.10} & 1.02 & 1.09 & 1.03 & \textbf{1.10} \\

flowers & 1.00 & 0.92 & \textbf{1.08} & 1.04 & 1.02 & \underline{1.07} & 1.00 & 1.02 & 0.95 & 0.98 \\

kitti\_closest\_vehicle\_distance & \textbf{1.00} & \underline{0.94} & 0.74 & 0.83 & 0.74 & 0.80 & 0.79 & 0.80 & 0.82 & 0.81 \\

pcam & \textbf{1.00} & \underline{1.00} & \underline{1.00} & \underline{1.00} & \underline{1.00} & \underline{1.00} & \underline{1.00} & \underline{1.00} & \underline{1.00} & \underline{1.00} \\

pets & 1.00 & 1.04 & \textbf{1.17} & \underline{1.16} & 1.14 & 1.14 & 1.13 & 1.11 & 0.99 & 0.99 \\

resisc45 & 1.00 & 0.75 & 0.80 & 0.79 & \underline{0.85} & 0.75 & 0.76 & \textbf{0.82} & 0.74 & 0.70 \\

smallnorb\_label\_azimuth & 1.00 & 1.13 & 1.14 & \underline{1.23} & 1.13 & \textbf{1.25} & 1.15 & 1.09 & 1.23 & 1.17 \\

smallnorb\_label\_elevation & 1.00 & 0.97 & 1.07 & 1.08 & 1.09 & 1.02 & \textbf{1.12} & \underline{1.10} & 1.07 & 1.05 \\

svhn & 1.00 & 1.04 & \textbf{1.51} & \underline{1.48} & 1.30 & 1.28 & 1.31 & 1.28 & 1.31 & 1.30 \\

\midrule
\textbf{Average Relative Accuracy} &
\textbf{1.00} & \textbf{0.87} &
\textbf{0.98} & \textbf{0.96} &
\textbf{0.96} & \textbf{0.95} &
\textbf{0.92} & \textbf{0.91} &
\textbf{0.91} & \textbf{0.92} \\
\bottomrule
\end{tabular}

\caption{Relative accuracy table with bold marking the best per row and underline marking the second-best per row.}
\label{tab:rel}
\end{table*}

\subsection{RGB Baselines: Additional Details}

\begin{figure*}[h!]
    \centering
    \begin{minipage}[c]{0.48\textwidth}
        \centering
        \includegraphics[width=\linewidth]{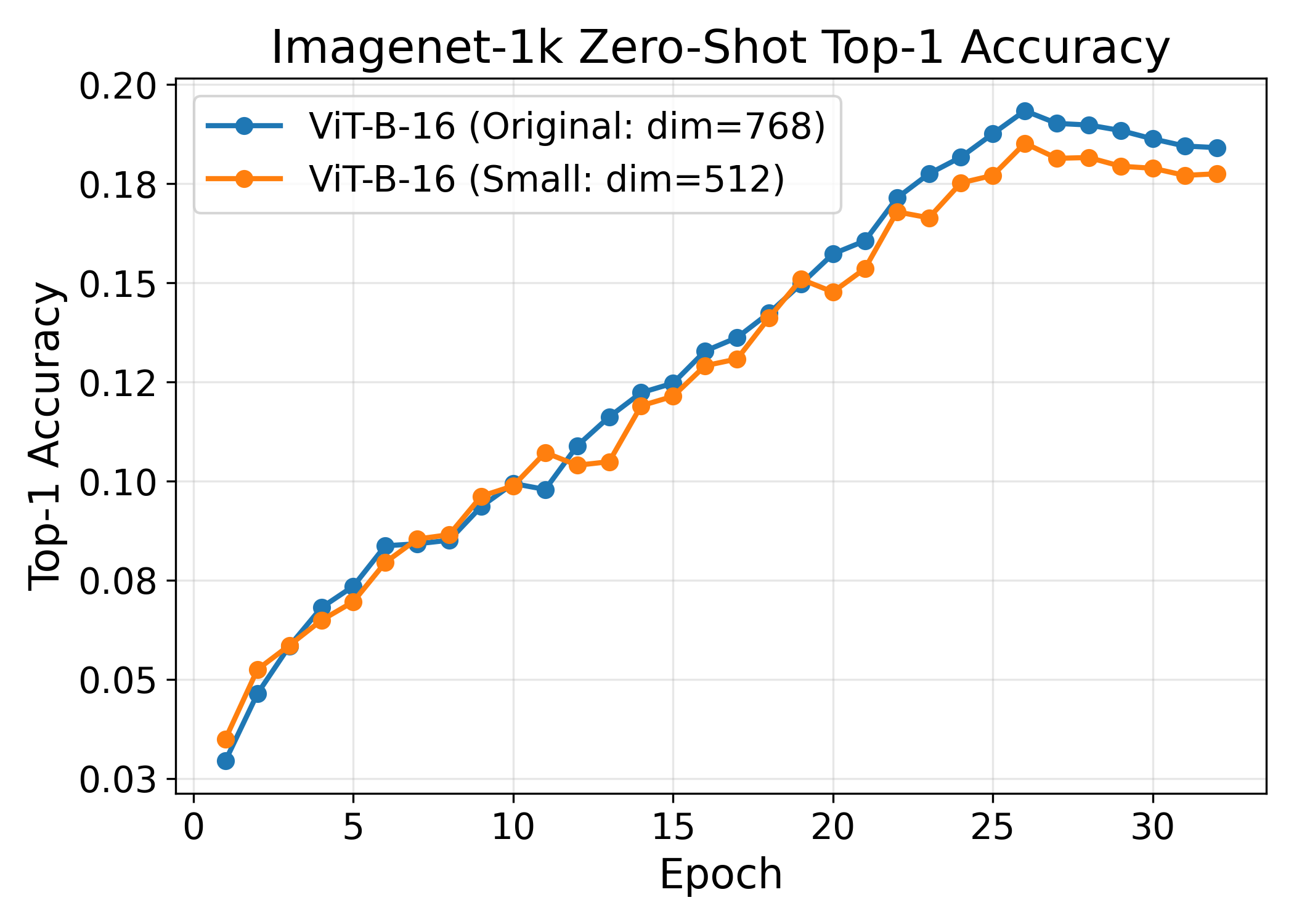}
    \end{minipage}
    \hfill
    \begin{minipage}[c]{0.48\textwidth}
        \centering
        \includegraphics[width=\linewidth]{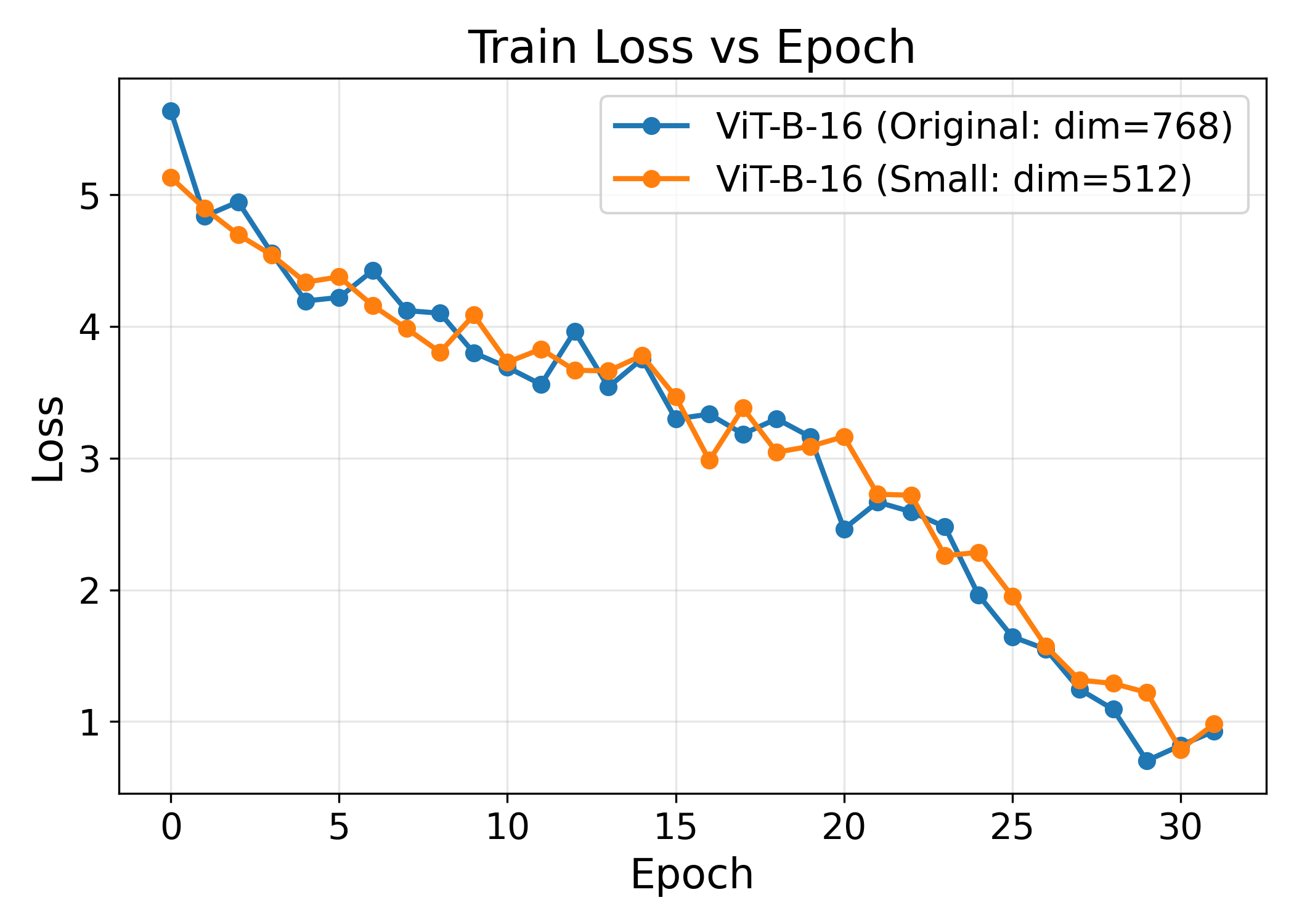}
    \end{minipage}

    \caption{
\textbf{Effect of ViT Width Reduction (768 $\rightarrow$ 512).}
Zero-shot accuracy (left) and train loss (right) for the standard ViT-B/16 (width 768) and our reduced-cost ViT-B/16 (Small) variant (width 512). 
Both models exhibit nearly identical behavior, validating the reduced-width encoder as an appropriate baseline.
}
    \label{vit_512}
\end{figure*}

\paragraph{ViT-B/16 (Small).}
To keep computational cost manageable during large-scale experimentation, we use a reduced-width ViT-B/16 variant whose hidden dimension is lowered from 768 to 512. 
Figure~\ref{vit_512} shows that this ``ViT-B/16 (Small)'' configuration maintains comparable training loss and zero-shot accuracy to the standard 768-width model, confirming that the reduced-width encoder is an appropriate baseline for our GS comparisons.

\paragraph{Token-Count Reduction for RGB Baselines.}
Throughout the main paper, the RGB ViT-B/16 encoder uses the standard 196-token patch embedding. 
For a fair comparison to our GS encoders, which natively support fewer latent primitives, we benchmarked 3 adaptations of ViT-B-16 (Small) operating with 98 tokens. 
This reduction is performed \emph{at the tokenization stage} by merging consecutive patch embeddings after positional encoding, rather than merging the full encoder output tokens.

\begin{figure*}[htbp]
  \centering
   \includegraphics[width=1\linewidth]{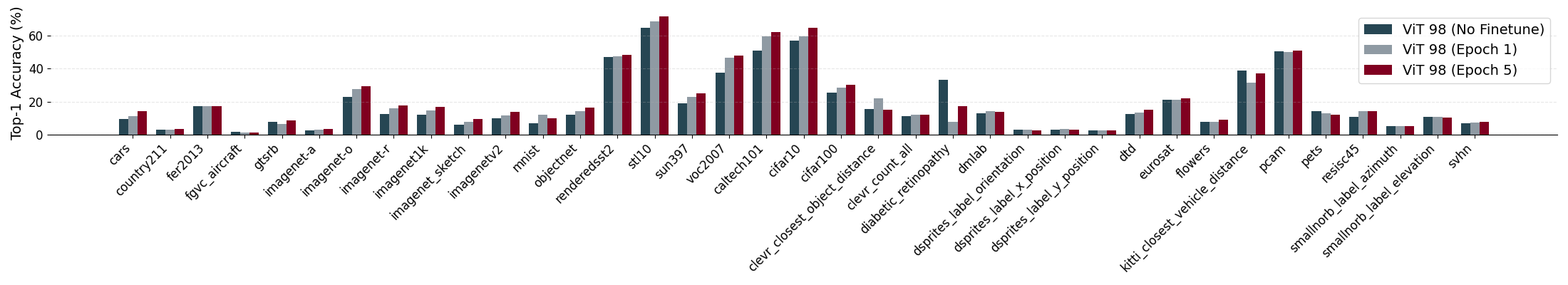}

   \caption{\textbf{Impact of Token-Count Reduction on RGB ViT Performance.}
Comparison of ViT-B/16 models using 98 tokens under no fine-tuning, 1-epoch fine-tuning, and 5-epoch fine-tuning.}
   \label{vit_98}
\end{figure*}

Because retraining a true 98-token ViT-B/16 from scratch (using larger patch sizes) is computationally expensive, we explore three adaptation settings: (1) no fine-tuning (direct token merging), (2) 1-epoch fine-tuning, and (3) 5-epoch fine-tuning, as shown in Table~\ref{tab:acc-vit98} and Fig.~\ref{vit_98}.
This illustrates that both RGB and GS tokenizations can both exhibit resilience to reduced token counts, with the highest 98 token relative accuracy achieved by the GS pipeline.

\begin{table*}[h!]
\centering
\small
\setlength{\tabcolsep}{5pt}
\renewcommand{\arraystretch}{1.08}

\begin{tabular}{lccc}
\toprule
\textbf{Dataset} & \textbf{ViT-98 no-ft} & \textbf{ViT-98 ep1} & \textbf{ViT-98 ep5} \\
\midrule
cars & 9.35 & 11.28 & 14.07 \\
country211 & 3.01 & 3.10 & 3.32 \\
fer2013 & 17.34 & 17.09 & 17.23 \\
fgvc\_aircraft & 1.59 & 1.17 & 1.32 \\
gtsrb & 7.81 & 6.50 & 8.67 \\
imagenet-a & 2.79 & 2.99 & 3.61 \\
imagenet-o & 22.95 & 27.80 & 29.40 \\
imagenet-r & 12.40 & 15.94 & 17.84 \\
imagenet1k & 12.14 & 14.47 & 16.64 \\
imagenet\_sketch & 6.06 & 7.86 & 9.49 \\
imagenetv2 & 9.88 & 11.69 & 13.64 \\
mnist & 6.97 & 11.99 & 10.13 \\
objectnet & 11.96 & 14.04 & 16.35 \\
renderedsst2 & 46.95 & 47.45 & 48.11 \\
stl10 & 64.79 & 68.66 & 71.43 \\
sun397 & 18.83 & 22.91 & 24.84 \\
voc2007 & 37.42 & 46.59 & 47.98 \\
caltech101 & 50.71 & 59.61 & 62.14 \\
cifar10 & 57.04 & 59.52 & 64.83 \\
cifar100 & 25.39 & 28.59 & 30.08 \\
clevr\_closest\_object\_distance & 15.58 & 21.90 & 15.20 \\
clevr\_count\_all & 11.27 & 12.29 & 11.92 \\
diabetic\_retinopathy & 33.21 & 7.59 & 17.40 \\
dmlab & 12.77 & 14.06 & 13.73 \\
dsprites\_label\_orientation & 3.01 & 2.96 & 2.56 \\
dsprites\_label\_x\_position & 2.86 & 3.43 & 3.05 \\
dsprites\_label\_y\_position & 2.60 & 2.43 & 2.55 \\
dtd & 12.55 & 13.24 & 14.89 \\
eurosat & 21.26 & 21.24 & 21.87 \\
flowers & 7.84 & 7.66 & 9.19 \\
kitti\_closest\_vehicle\_distance & 38.82 & 31.65 & 37.27 \\
pcam & 50.30 & 50.17 & 50.76 \\
pets & 14.17 & 12.92 & 11.94 \\
resisc45 & 10.87 & 14.10 & 14.19 \\
smallnorb\_label\_azimuth & 5.14 & 5.00 & 5.00 \\
smallnorb\_label\_elevation & 10.78 & 10.91 & 10.40 \\
svhn & 6.87 & 7.36 & 7.58 \\
\midrule 
\textbf{Average Absolute Accuracy} & \textbf{18.52} & \textbf{19.41} &	\textbf{20.56} \\
\midrule 
\textbf{Avg Relative Accuracy to Base (ViT 196)	} & \textbf{0.87} & \textbf{0.91} &	\textbf{0.95}\\ 
\bottomrule

\end{tabular}

\caption{Top-1 accuracy (\%) of ViT-B/16 with 98 tokens under three settings: no fine-tuning, 1-epoch fine-tuning, and 5-epoch fine-tuning.}
\label{tab:acc-vit98}
\end{table*}

\subsection{Model Architecture and Training Recipe Studies}
Across the course of this work, we conducted extensive experimental sweeps to evaluate architectural choices, training recipes, and design decisions that influence the representational capabilities of 2D Gaussian Splats. 
This section summarizes a focused subset of these studies, highlighting the settings that most strongly affected performance and optimization behavior.

\begin{figure*}[h!]
    \centering
    \subfloat[Zero-shot Top-1 after 1 Epoch]{
        \includegraphics[width=0.48\linewidth]{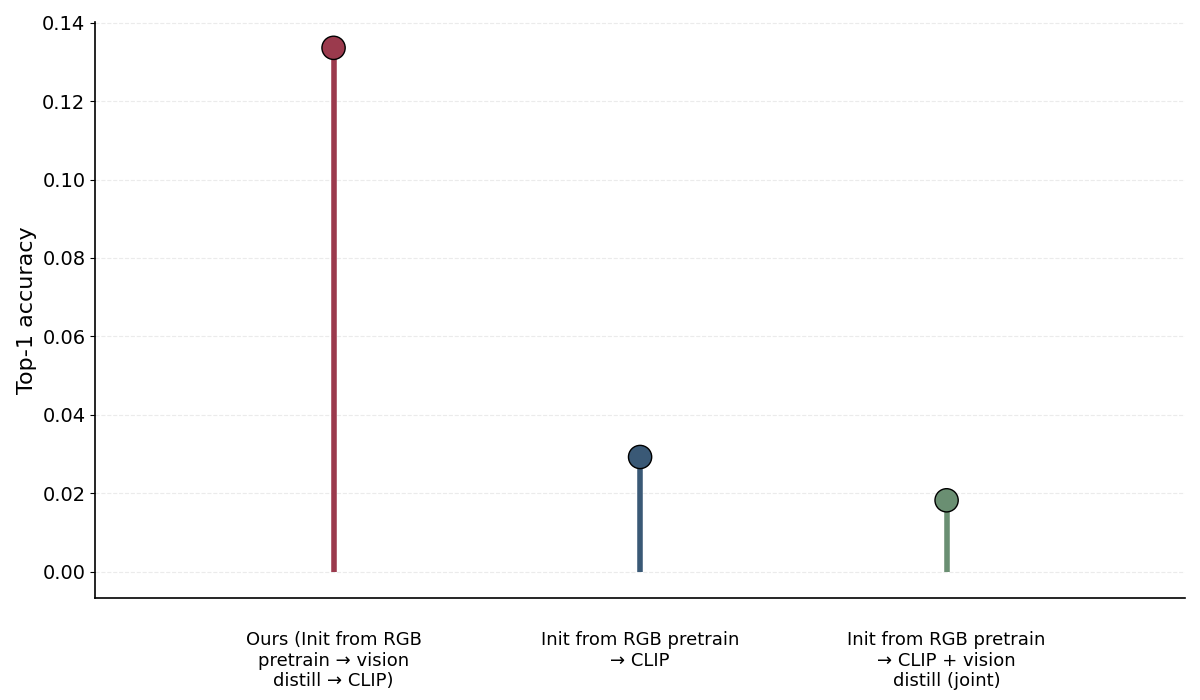}
    }
    \hfill
    \subfloat[Training Loss]{
        \includegraphics[width=0.48\linewidth]{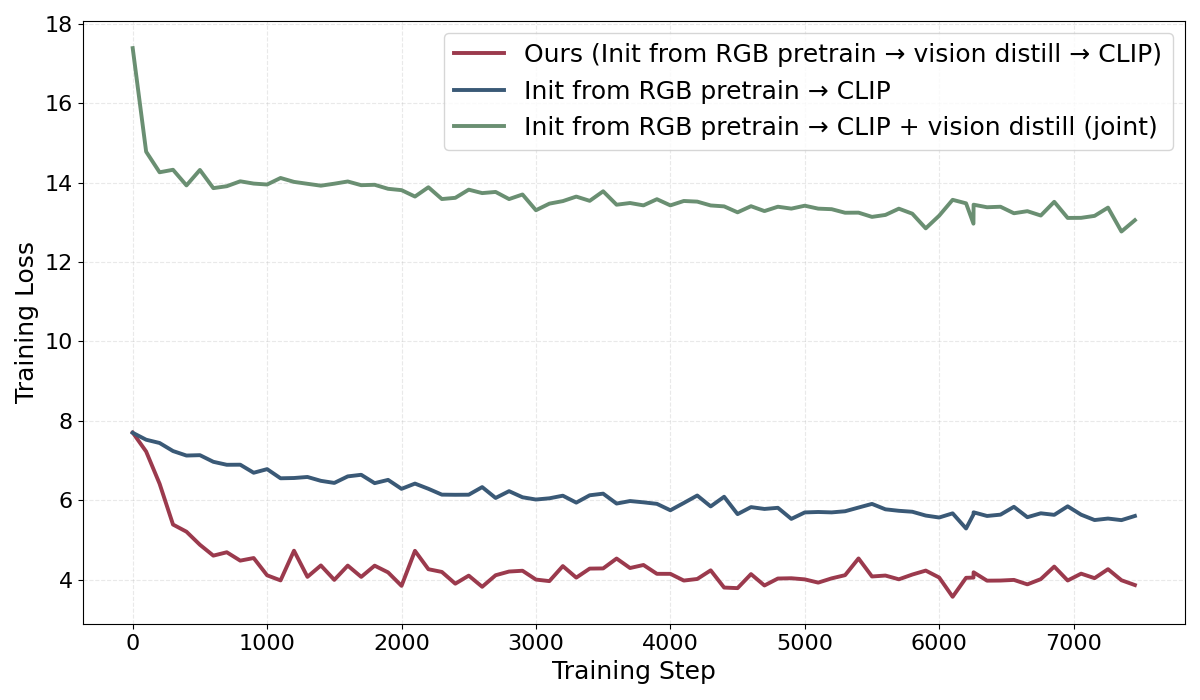}
    }
    \caption{Comparison of distillation strategies and their effect on CLIP alignment
    and optimization dynamics.}
    \label{fig:distill_compare}
\end{figure*}

\paragraph{Distillation vs. No-Distillation Ablations.}
Figure~\ref{fig:distill_compare} compares three initialization and supervision strategies: (1) RGB-pretrained ViT initialization, (2) vision-only distillation from a frozen RGB teacher followed by CLIP adaptation, (3) Simultaneous CLIP+\!vision-distillation.
Our approach of \textbf{first} performing vision-distillation, \textbf{and then} CLIP-adaptation largely outperforms both other techniques, and makes the best use of both vision-vision and text-vision supervision.

\begin{figure*}[h!]
    \centering
    \hspace{2cm}
    \subfloat[Zero-shot Top-1 after 1 Epoch]{
        \includegraphics[width=0.3\linewidth]{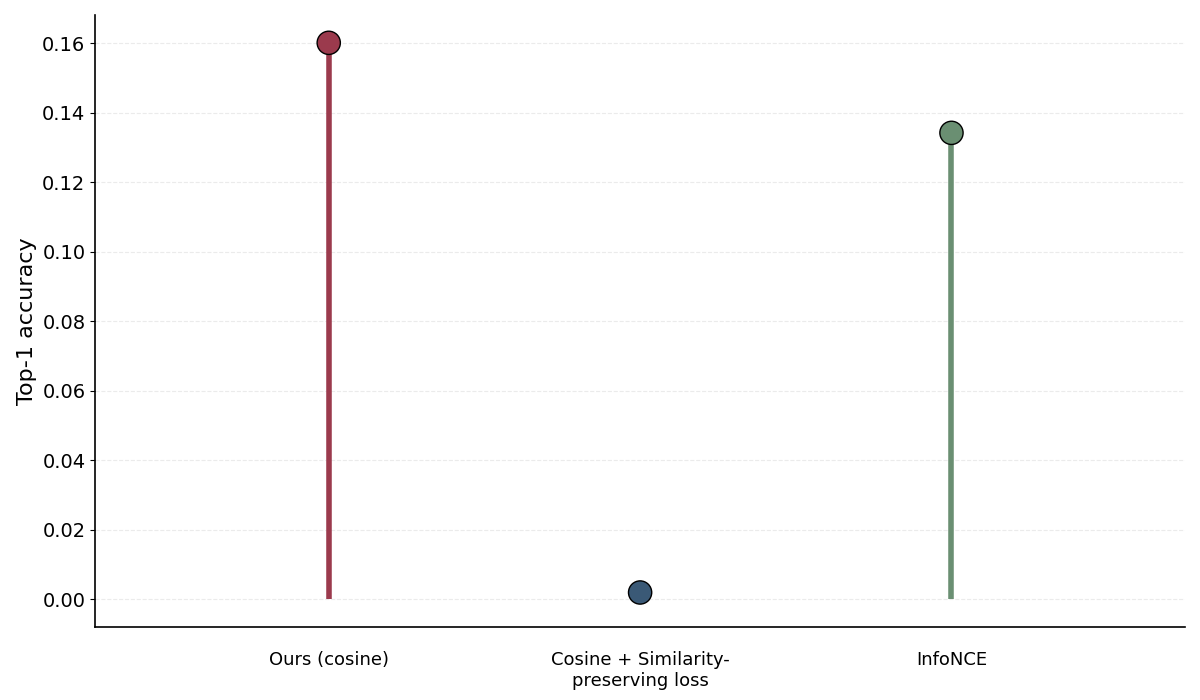}
    }
    \hspace{0.5cm}
    \subfloat[Training Loss]{
        \includegraphics[width=0.48\linewidth]{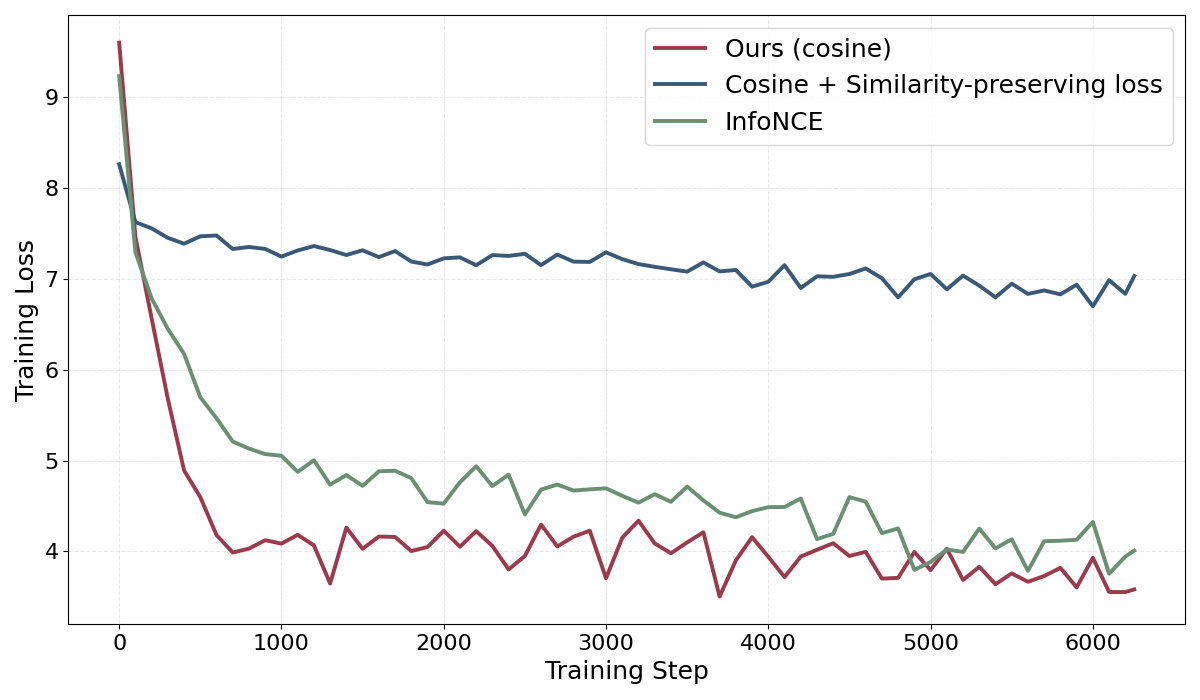}
    }
    \caption{Comparison of distillation losses and their effect on CLIP alignment
    and optimization dynamics.}
    \label{fig:distillloss}
\end{figure*}

\paragraph{Distillation Loss Variants.}
We evaluated multiple distillation objectives, including cosine similarity, InfoNCE, and cosine similarity augmented with a similarity-preserving constraint inspired by \cite{tian2019contrastive}.
For the latter, we applied a layerwise similarity loss with weight $\gamma=2000$ to encourage structural alignment between the GS and RGB feature spaces.
Figure~\ref{fig:distillloss} shows that although similarity-preserving and InfoNCE losses improve stability, simple cosine-matching achieves the highest CLIP-level zero-shot accuracy and offers the most reliable convergence behavior in practice.

\begin{figure*}[h!]
    \centering
    \hspace{2cm}
    \subfloat[Zero-shot Top-1]{
        \includegraphics[width=0.3\linewidth]{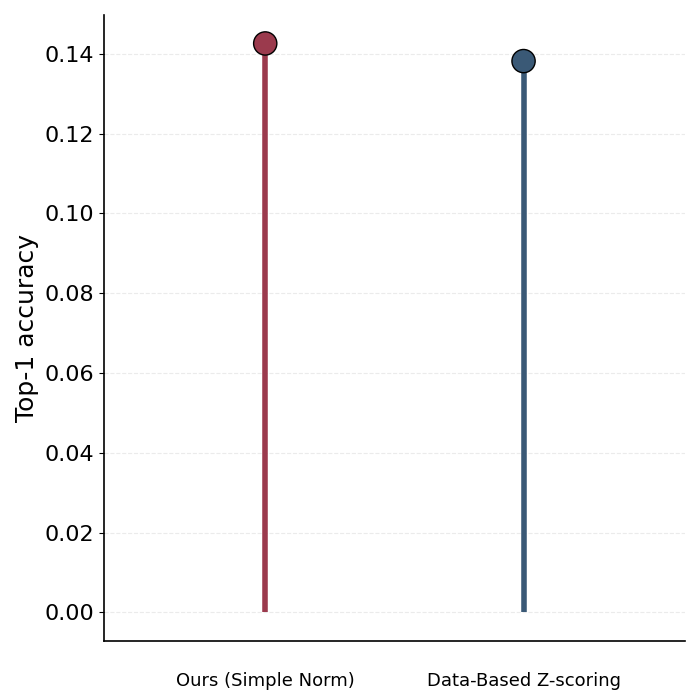}
    }
    \hspace{0.5cm}
    \subfloat[Training Loss]{
        \includegraphics[width=0.48\linewidth]{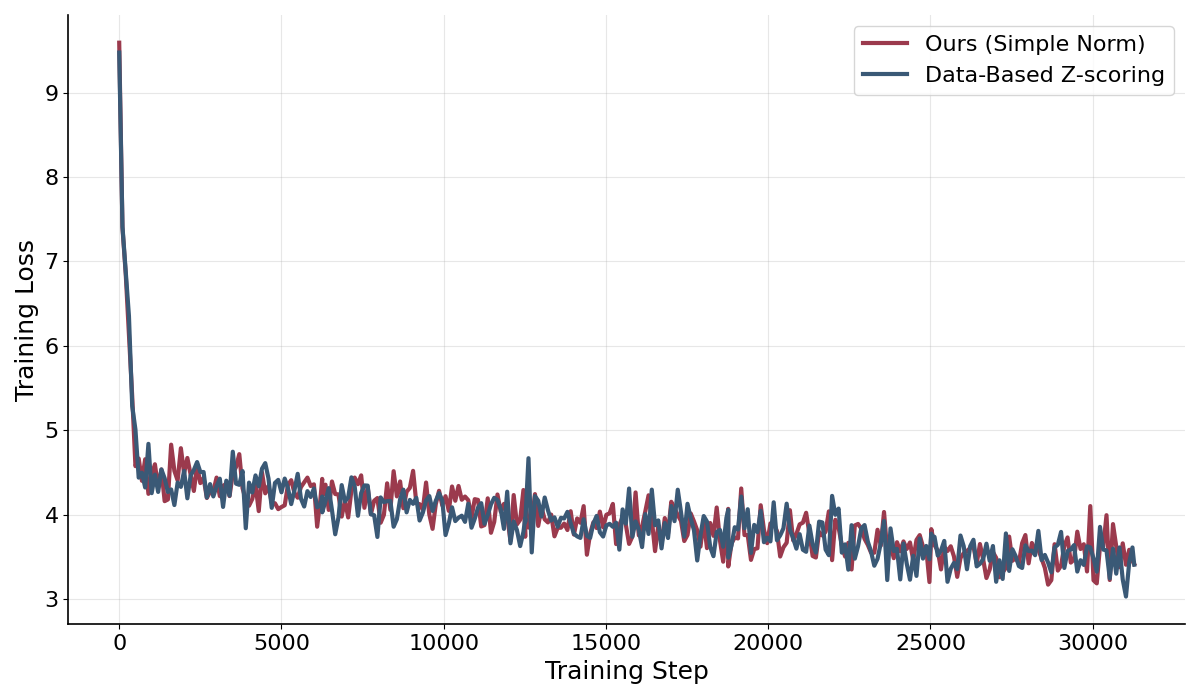}
    } \\[1em]

    \caption{Normalization study: effects of dataset-level normalization vs. simple rescaling.}
    \label{fig:norm_study}
\end{figure*}

\paragraph{Normalization Studies.}
We additionally assessed the effect of normalization choices on CLIP training dynamics (Fig.~\ref{fig:norm_study}).
We found log-scaling of covariance components as well as normalization of the x,y components by the resolution to work better than z-score normalization using the collected dataset statistics.

\begin{figure*}[h!]
    \centering
    \subfloat[Zero-shot Top-1 after 1 Epoch]{
        \includegraphics[width=0.48\linewidth]{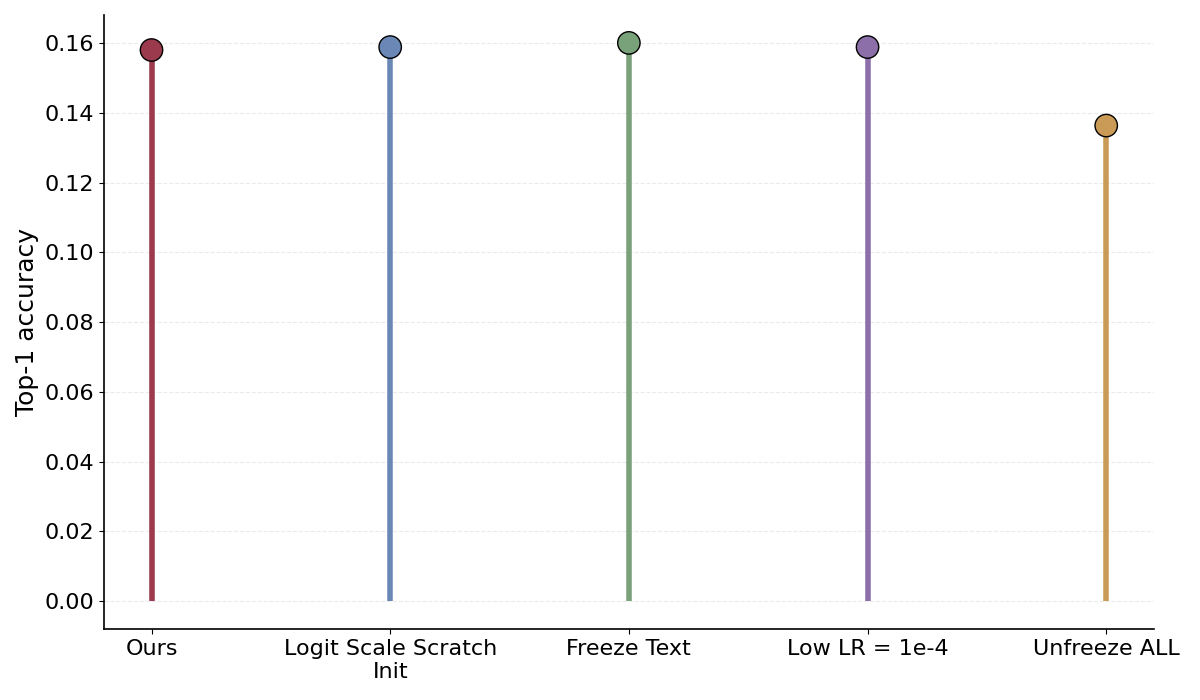}
    }
    \hfill
    \subfloat[Training Loss]{
        \includegraphics[width=0.48\linewidth]{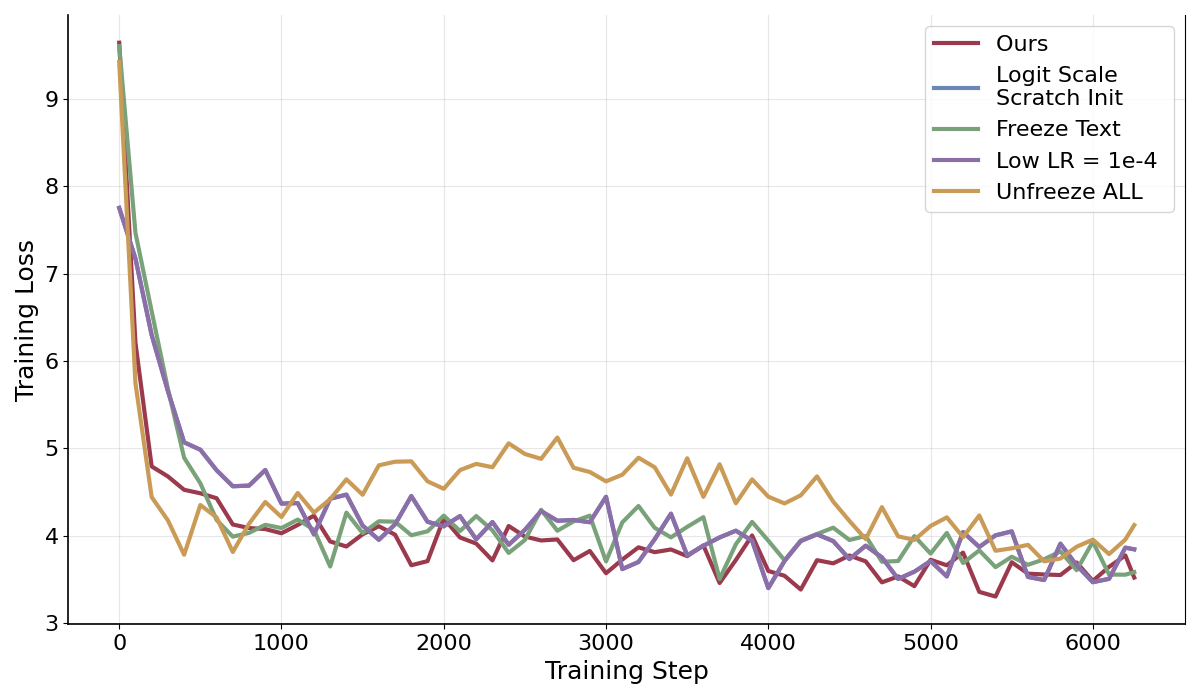}
    }
    \caption{Hyperparameter sensitivity analysis across logit-scale initialization,
    learning rate, and text-freezing behaviors.}
    \label{fig:hyperparams_study}
\end{figure*}

\paragraph{Hyperparameter Ablations.}
Figure~\ref{fig:hyperparams_study} presents ablations over logit-scale initialization, text-encoder freezing policies, and learning-rate magnitude.
Logit-scale transfer from the RGB baseline improves early performance, whereas fully freezing the text tower degrades convergence speed in later epochs.

\begin{figure*}[h!]
    \centering

    \subfloat[Zero-shot Top-1 after 1 Epoch]{
        \includegraphics[width=0.48\linewidth]{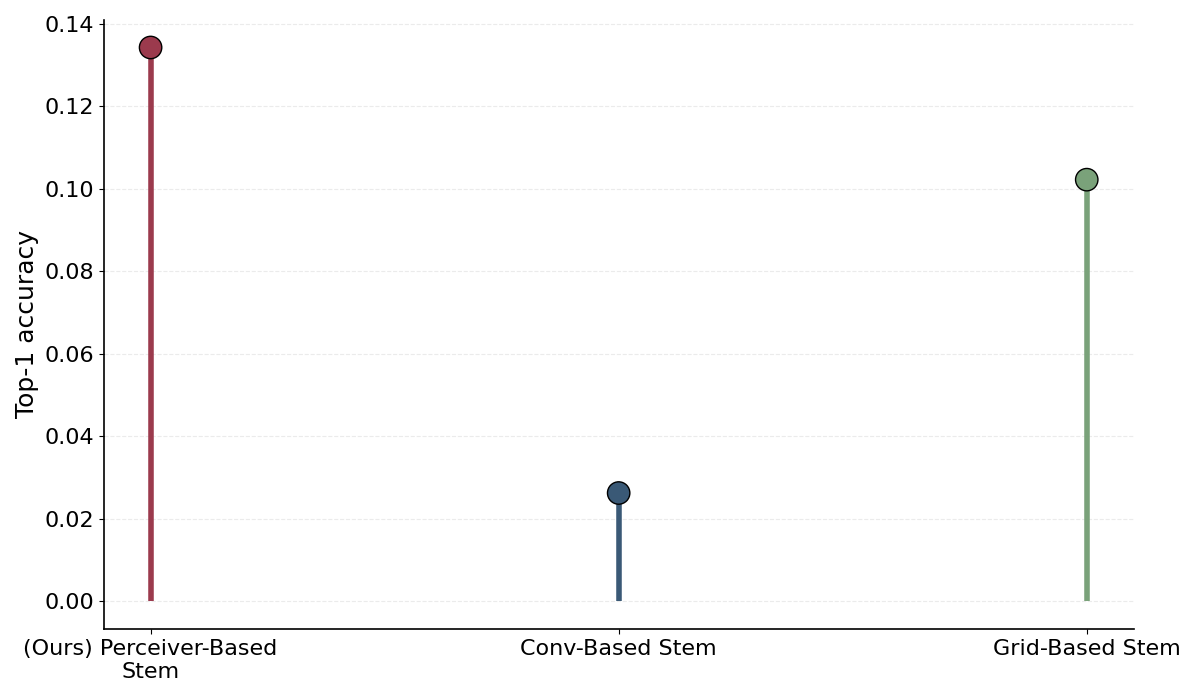}
    }
    \hfill
    \subfloat[Training Loss]{
        \includegraphics[width=0.48\linewidth]{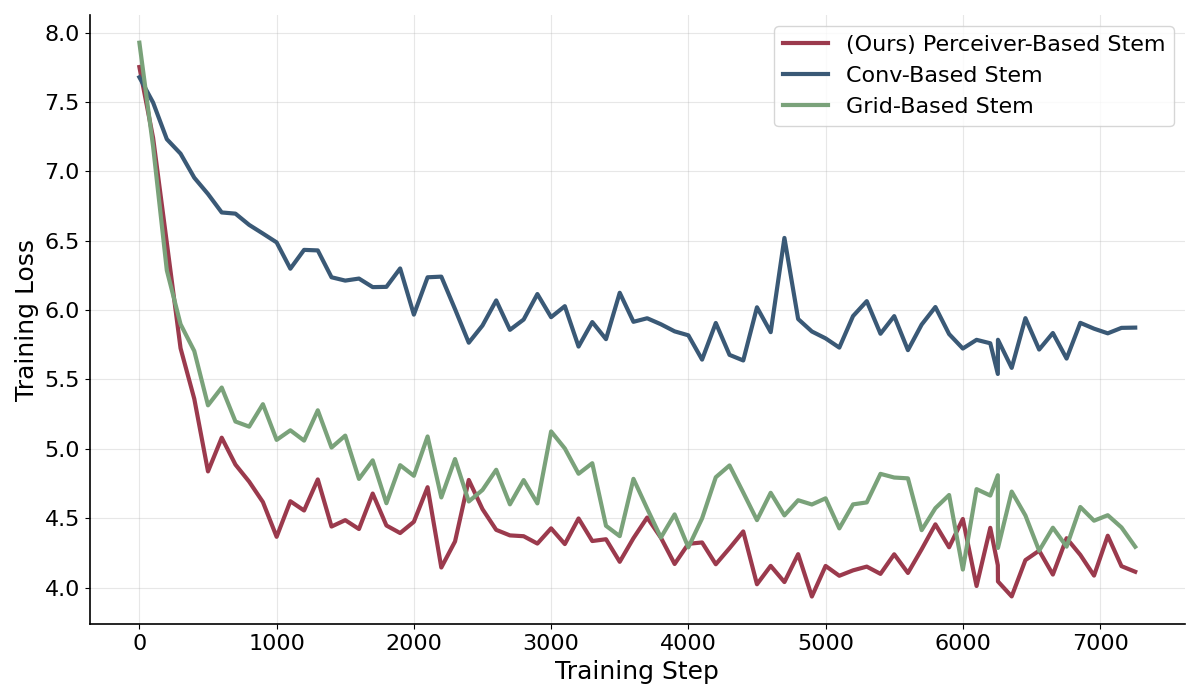}
    }
    \caption{GS-Stem architecture study: different approaches to going from N points to smaller M latents.}
    \label{fig:archs}
\end{figure*}

\paragraph{GSStem Architectures.}
We explored several designs for projecting $N$ Gaussian points into a compact set of $M$ latent tokens. 
As shown in Fig.~\ref{fig:archs}, a Perceiver-style cross-attention stem produces the best zero-shot performance and the smoothest training loss.
Alternative stems, including (i) grid-based Gaussian pooling that mimics ViT patchification and (ii) Hilbert-ordered chunking with localized convolution over point sequences, proved less effective.
The Perceiver design consistently extracted richer, more expressive latent tokens.

\begin{figure*}[h!]
    \centering
    \hspace{2cm}
    \subfloat[Zero-shot Top-1 after 1 Epoch]{
        \includegraphics[width=0.3\linewidth]{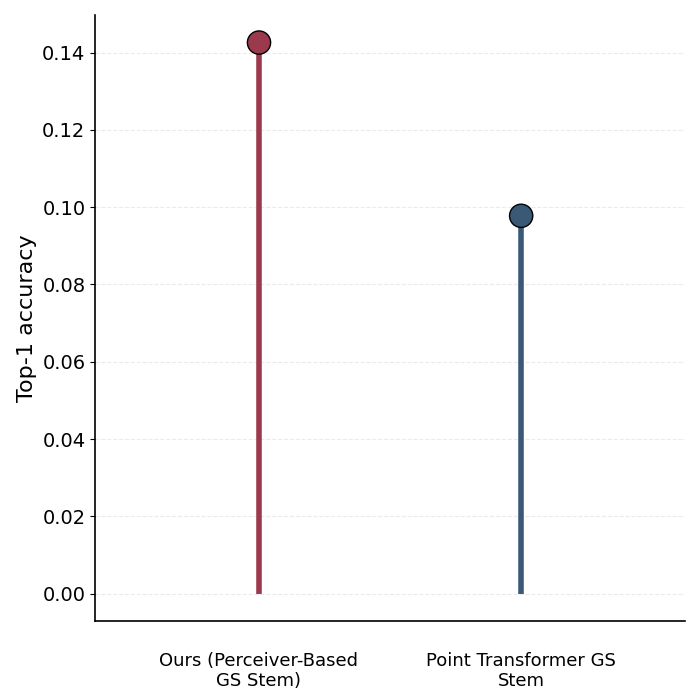}
    }
    \hspace{0.5cm}
    \subfloat[Training Loss]{
        \includegraphics[width=0.48\linewidth]{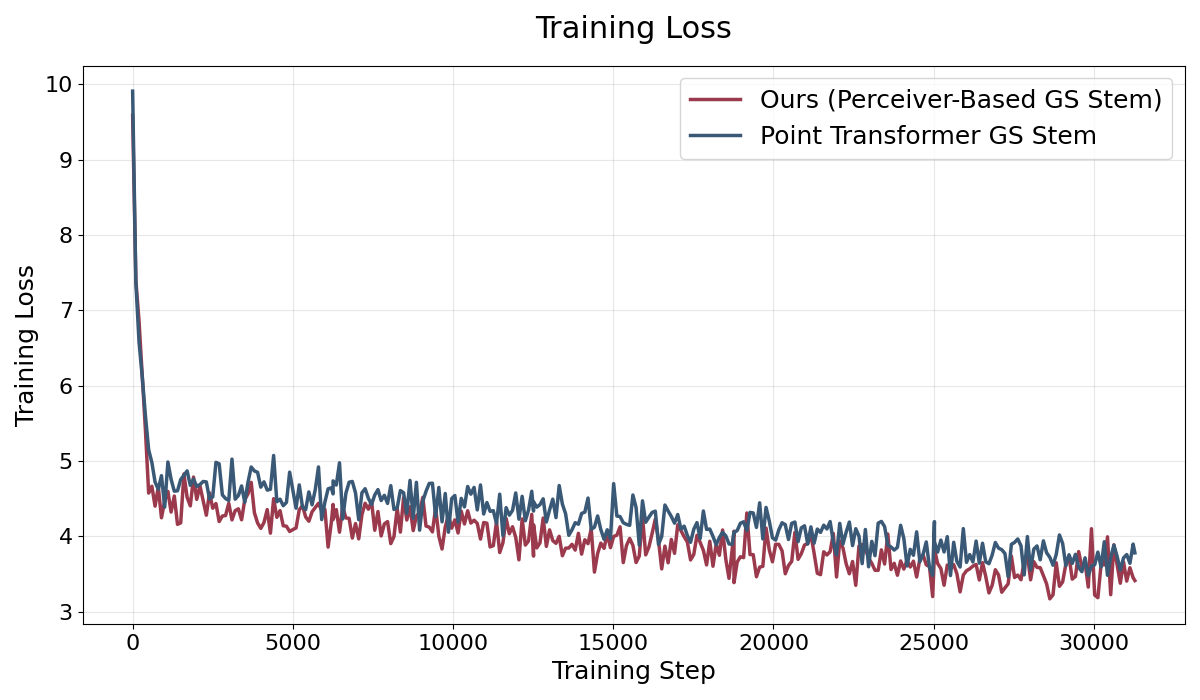}
    }
    \caption{Point Transformer ablation: early-stage CLIP alignment performance when varying initialization and distillation strategy.}
    \label{fig:pt_study}
\end{figure*}

\paragraph{Point Transformer Study.}
Inspired by the 3D origin of Gaussian-splatting, we also evaluated point-based transformer encoders for 2DGS features.
Figure~\ref{fig:pt_study} shows zero-shot accuracy after one epoch of training (left) and the corresponding training losses (right). We found point transformer based models to be less stable to train and suboptimal in terms of performance compared to perceiver architectures, albeit point transformers’ ability to emit variable token counts aligns well with 2DGS adaptivity, and merits further study.

\subsection{Case Studies on Representational Power and Bottlenecks of 2DGS.}
To better understand the fundamental capabilities and limitations of 2D Gaussian Splat (2DGS) representations, we conducted a series of controlled case studies, summarized in Figs.~14--16. These analyses isolate the effects of lossy compression, lack of pretrained inductive bias, and architectural choices on downstream representational performance.

\begin{enumerate}[leftmargin=1.2em]

\item \textbf{Rendering Back Into Pixels.}
We render our GS-1600 dataset back into RGB pixel space and retrain a ViT-B/16 (Small) encoder on the reconstructed images (Fig.~14). 
As expected for a lossy compressor, accuracy declines relative to training on original RGB, but the \emph{relative drop} closely mirrors the performance drop observed when training CLIP using GS representations directly. 
This alignment suggests that our Perceiver-based GS encoder is extracting the majority of the semantic content available in the rendered images, and that the overall performance ceiling is fundamentally constrained by the quality of the compressed RGB surrogate. 
Because our GS encoder is distilled from an RGB-pretrained teacher, its achievable zero-shot performance is similarly bounded by the representational limits of the underlying pixel-domain ViT.

\item \textbf{Training GS Encoders Fully From Scratch.}
We train our GS-1600 (196-token) encoder \emph{entirely from scratch}, without RGB initialization, pretraining, or distillation (Fig.~15). 
This model exhibits substantially lower zero-shot accuracy and slower convergence, highlighting the absence of strong native inductive biases for 2DGS representations when trained without external guidance. 
In contrast, distillation from RGB features provides a powerful initialization signal that lifts the GS encoder out of the suboptimal optimization basin associated with scratch training, emphasizing the importance of cross-domain supervision for early-stage 2DGS models.

\item \textbf{2DGS as a Tokenizer for Vanilla Transformers.}
To test the representational power of 2DGS in extreme compression regimes without specialized architectures, we feed a low number of Gaussian splats directly into a vanilla Transformer after lightweight per-channel normalization, rescaling, and Fourier feature augmentation (Fig.~16). We pre-train the GS encoder from scratch in this case.
We evaluate extreme compression settings (196 points) and low ones (400 points). 
Accuracy lags behind architectures that explicitly map from $N$ splats to $M$ latents (e.g., Perceiver stems). 
The results indicate that starting from a richer Gaussian budget yields better fidelity and that an architectural bottleneck is required to appropriately condense spatially distributed Gaussians into semantically meaningful tokens. 
This points toward the need for future architectures that can \emph{adaptively} produce the number of output tokens based on the density and structure of the input splats.

\end{enumerate}


\begin{figure*}[h!]
    \centering
    \begin{minipage}[c]{0.48\textwidth}
        \centering
        \includegraphics[width=\linewidth]{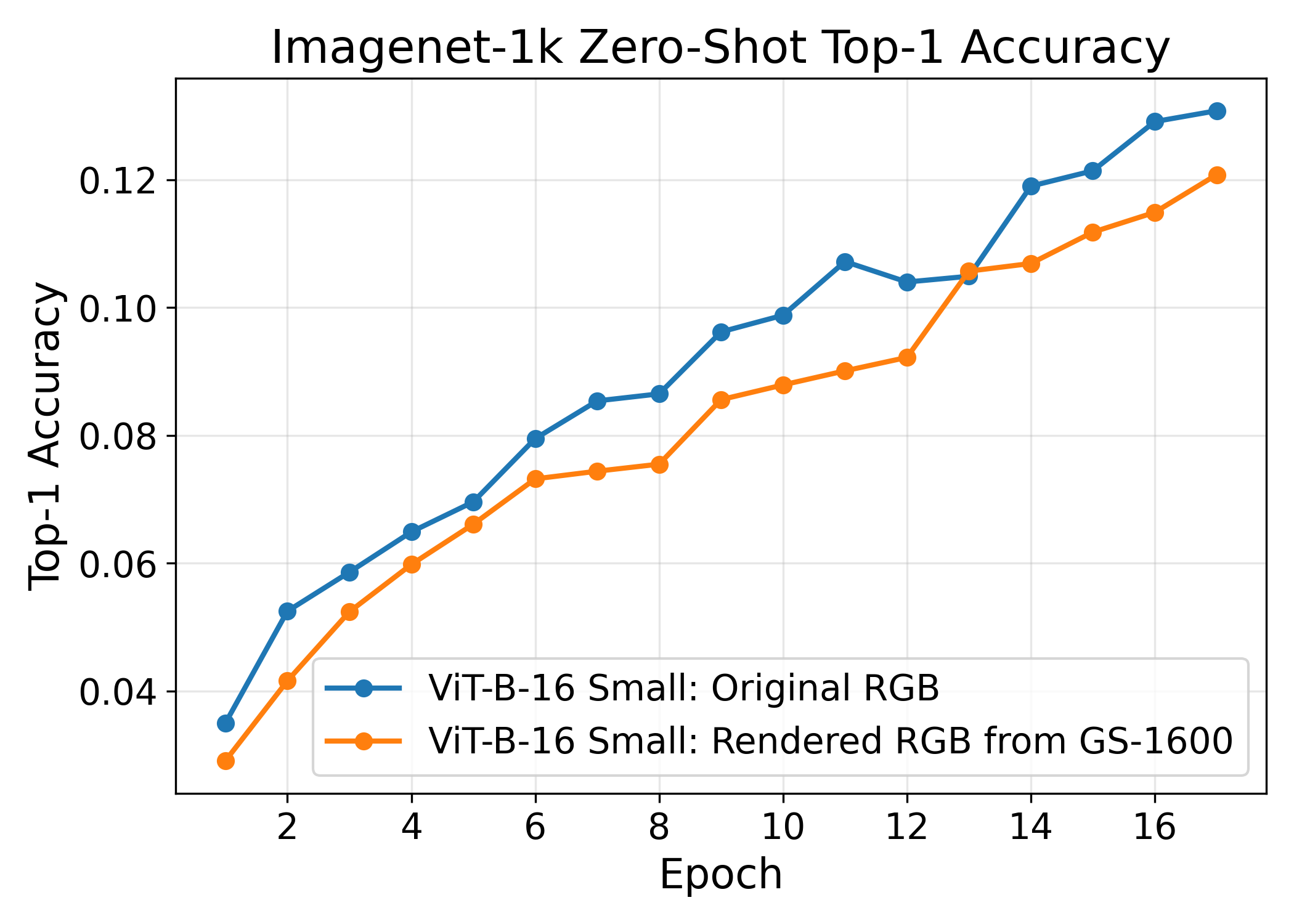}
    \end{minipage}
    \hfill
    \begin{minipage}[c]{0.48\textwidth}
        \centering
        \includegraphics[width=\linewidth]{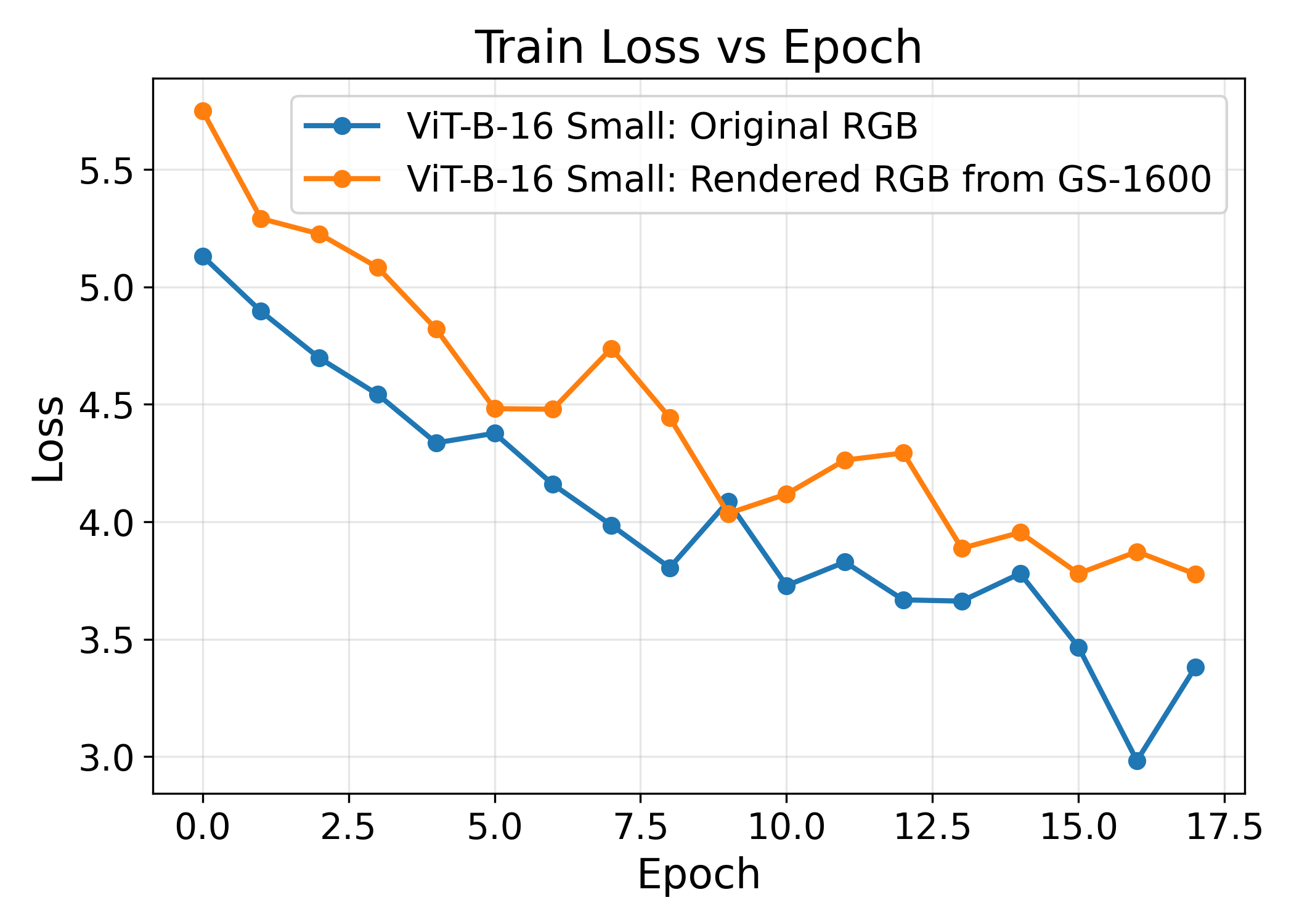}
    \end{minipage}
    \caption{\textbf{Rendered ViT:} Zero-shot accuracy (left) and train loss (right).}
\end{figure*}

\begin{figure*}[h!]
    \centering
    \begin{minipage}[c]{0.48\textwidth}
        \centering
        \includegraphics[width=\linewidth]{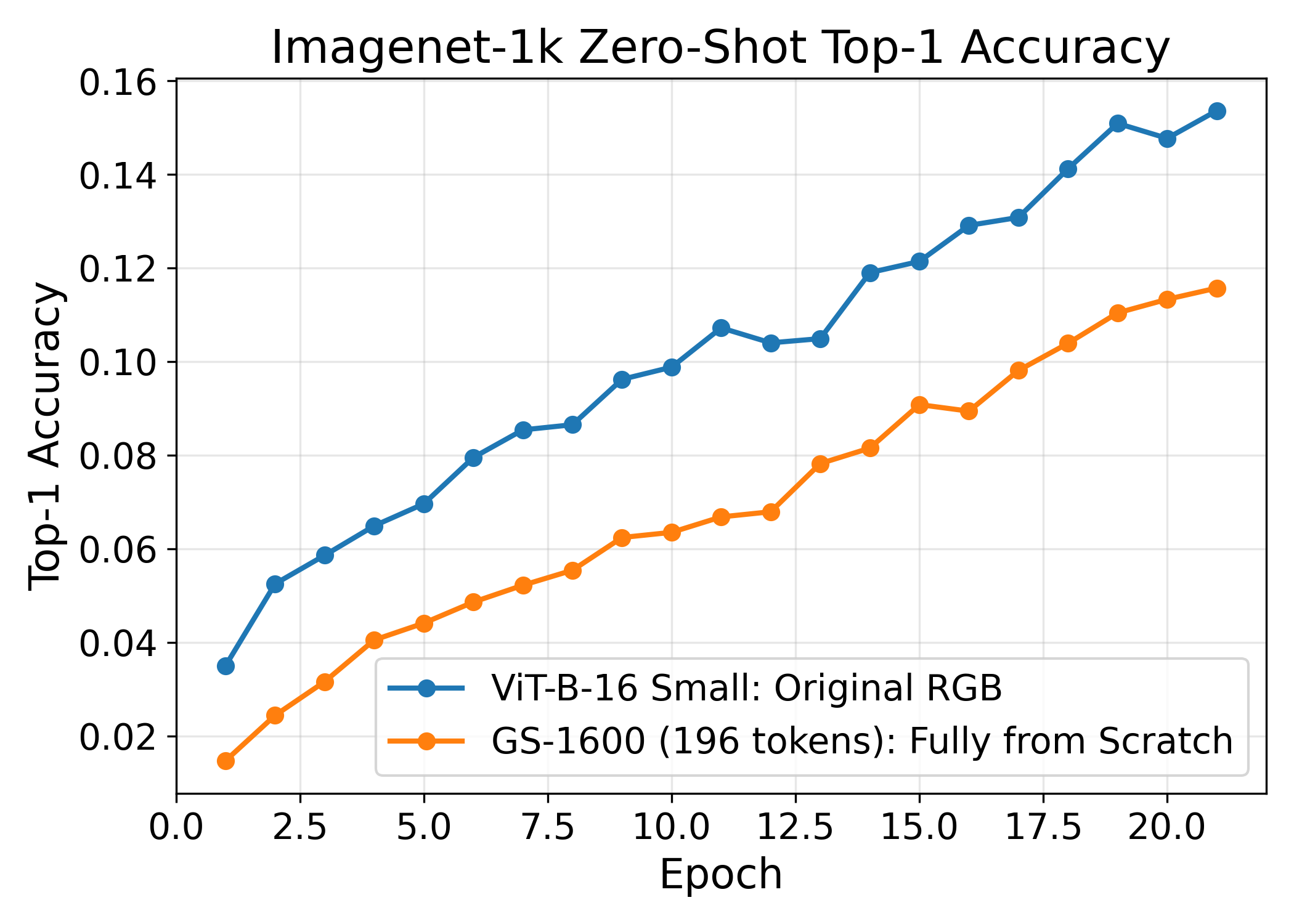}
    \end{minipage}
    \hfill
    \begin{minipage}[c]{0.48\textwidth}
        \centering
        \includegraphics[width=\linewidth]{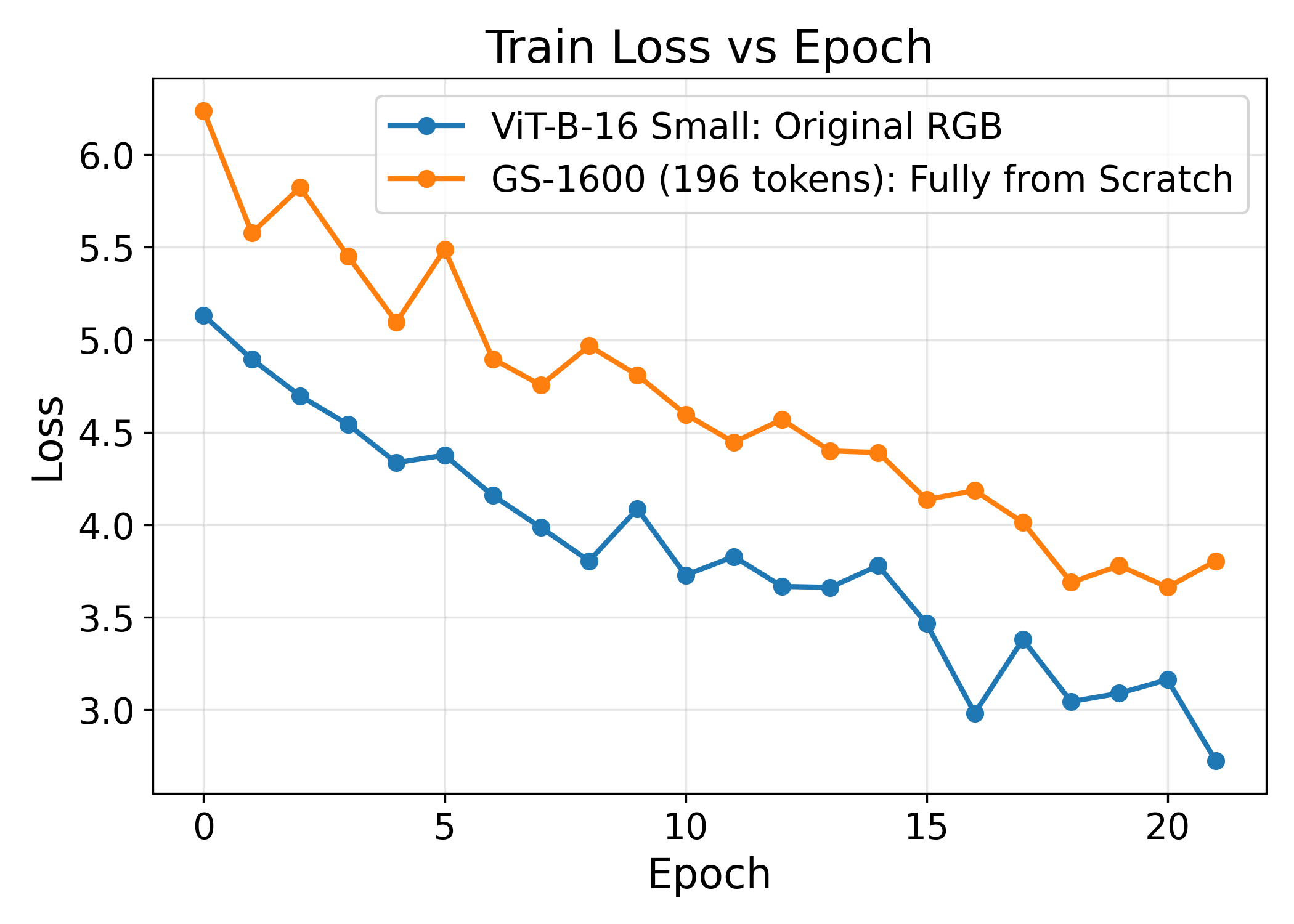}
    \end{minipage}
    \caption{\textbf{Training From Scratch:} Zero-shot accuracy (left) and train loss (right).}
\end{figure*}

\begin{figure*}[h!]
    \centering
    \begin{minipage}[c]{0.48\textwidth}
        \centering
        \includegraphics[width=\linewidth]{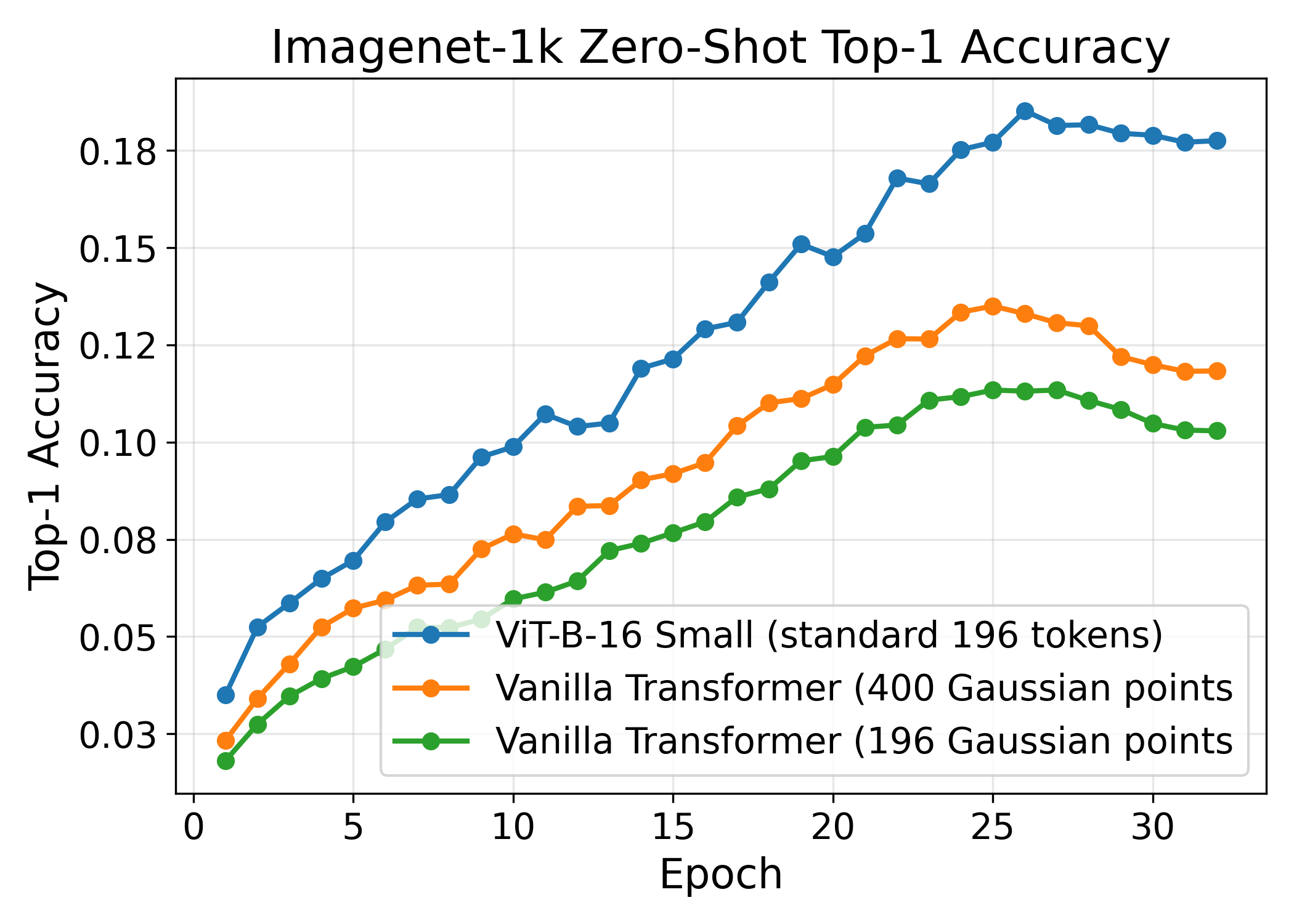}
    \end{minipage}
    \hfill
    \begin{minipage}[c]{0.48\textwidth}
        \centering
        \includegraphics[width=\linewidth]{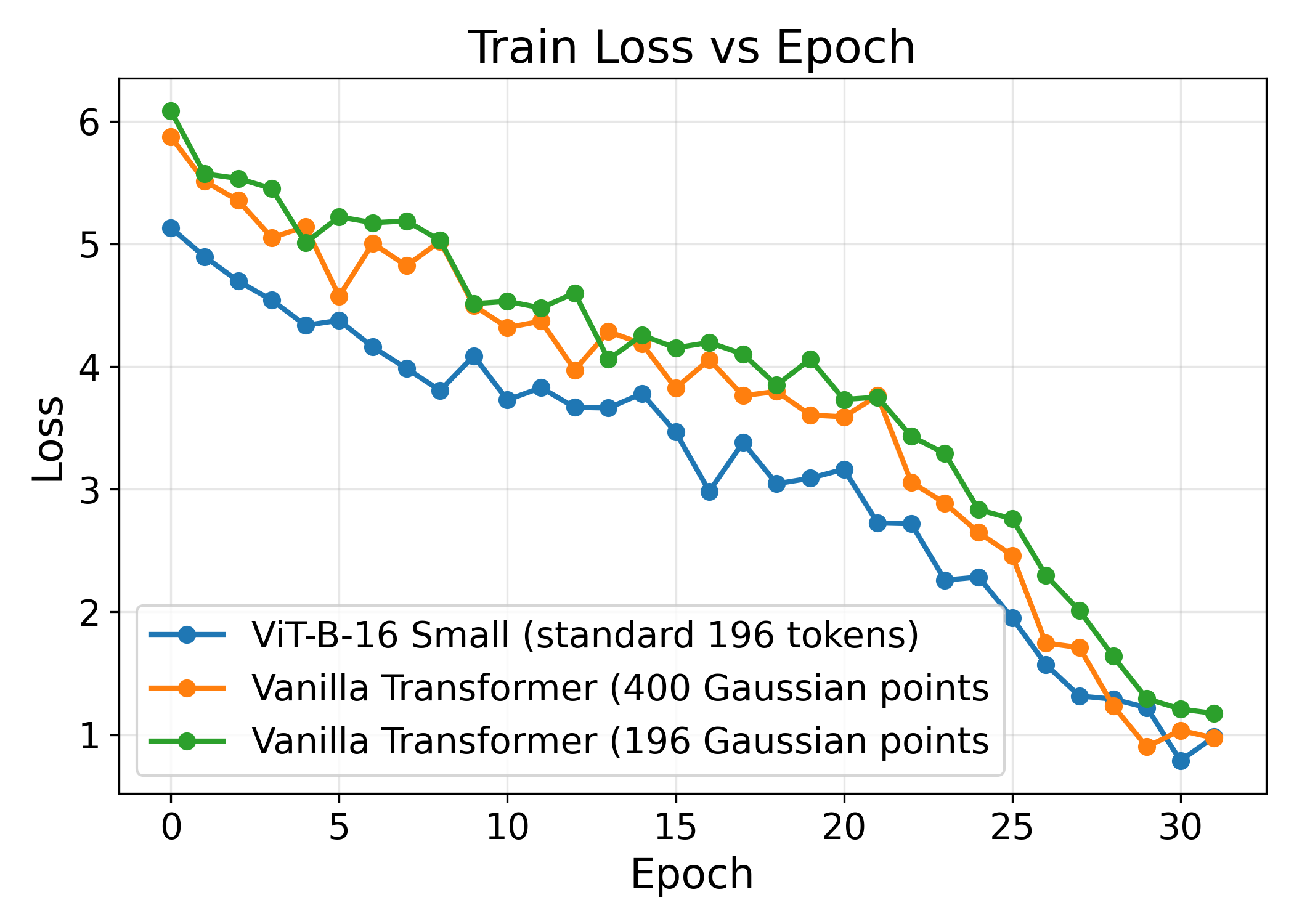}
    \end{minipage}
    \caption{\textbf{Token Ablation Study:} Zero-shot accuracy (left) and train loss (right).}
\end{figure*}





\end{document}